\documentclass[letterpaper]{article} 
\usepackage{aaai2026}  
\usepackage{times}  
\usepackage{helvet}  
\usepackage{courier}  
\usepackage[hyphens]{url}  
\usepackage{graphicx} 
\usepackage{natbib}  
\usepackage{caption} 
\usepackage{algorithm}
\usepackage{algorithmic}
\usepackage{amsmath}
\usepackage{amsfonts}       

\usepackage{newfloat}
\usepackage{listings}
\DeclareCaptionStyle{ruled}{labelfont=normalfont,labelsep=colon,strut=off} 
\floatstyle{ruled}
\newfloat{listing}{tb}{lst}{}
\floatname{listing}{Listing}
\usepackage{booktabs} 
\usepackage{multirow} 
\usepackage[table]{xcolor}
\usepackage{tabularx}

\title{VGLD: \underline{V}isually-\underline{G}uided \underline{L}inguistic \underline{D}isambiguation for Monocular Depth Scale Recovery}
\author{
    Bojin Wu, Jing Chen\thanks{corresponding author}\\
}
\affiliations{
    School of Physics and Optoelectronic Engineering,\\

    Guangdong University of Technology\\
    Guangzhou, 510000, China\\
    realpakinwu@gmail.com , jchen125@gdut.edu.cn
}

\usepackage{bibentry}

\begin{document}

\maketitle

\begin{abstract}
Monocular depth estimation can be broadly categorized into two directions: relative depth estimation, which predicts normalized or inverse depth without absolute scale, and metric depth estimation, which aims to recover depth with real-world scale. While relative methods are flexible and data-efficient, their lack of metric scale limits their utility in downstream tasks. A promising solution is to infer absolute scale from textual descriptions. However, such language-based recovery is highly sensitive to natural language ambiguity, as the same image may be described differently across perspectives and styles.
To address this, we introduce VGLD (Visually-Guided Linguistic Disambiguation), a framework that incorporates high-level visual semantics to resolve ambiguity in textual inputs. By jointly encoding both image and text, VGLD predicts a set of global linear transformation parameters that align relative depth maps with metric scale. This visually grounded disambiguation improves the stability and accuracy of scale estimation.
We evaluate VGLD on representative models, including MiDaS and DepthAnything, using standard indoor (NYUv2) and outdoor (KITTI) benchmarks. Results show that VGLD significantly mitigates scale estimation bias caused by inconsistent or ambiguous language, achieving robust and accurate metric predictions. Moreover, when trained on multiple datasets, VGLD functions as a universal and lightweight alignment module, maintaining strong performance even in zero-shot settings. Code will be released upon acceptance.
\end{abstract}


\begin{figure}[!ht]
\centering
\includegraphics[width=\columnwidth]{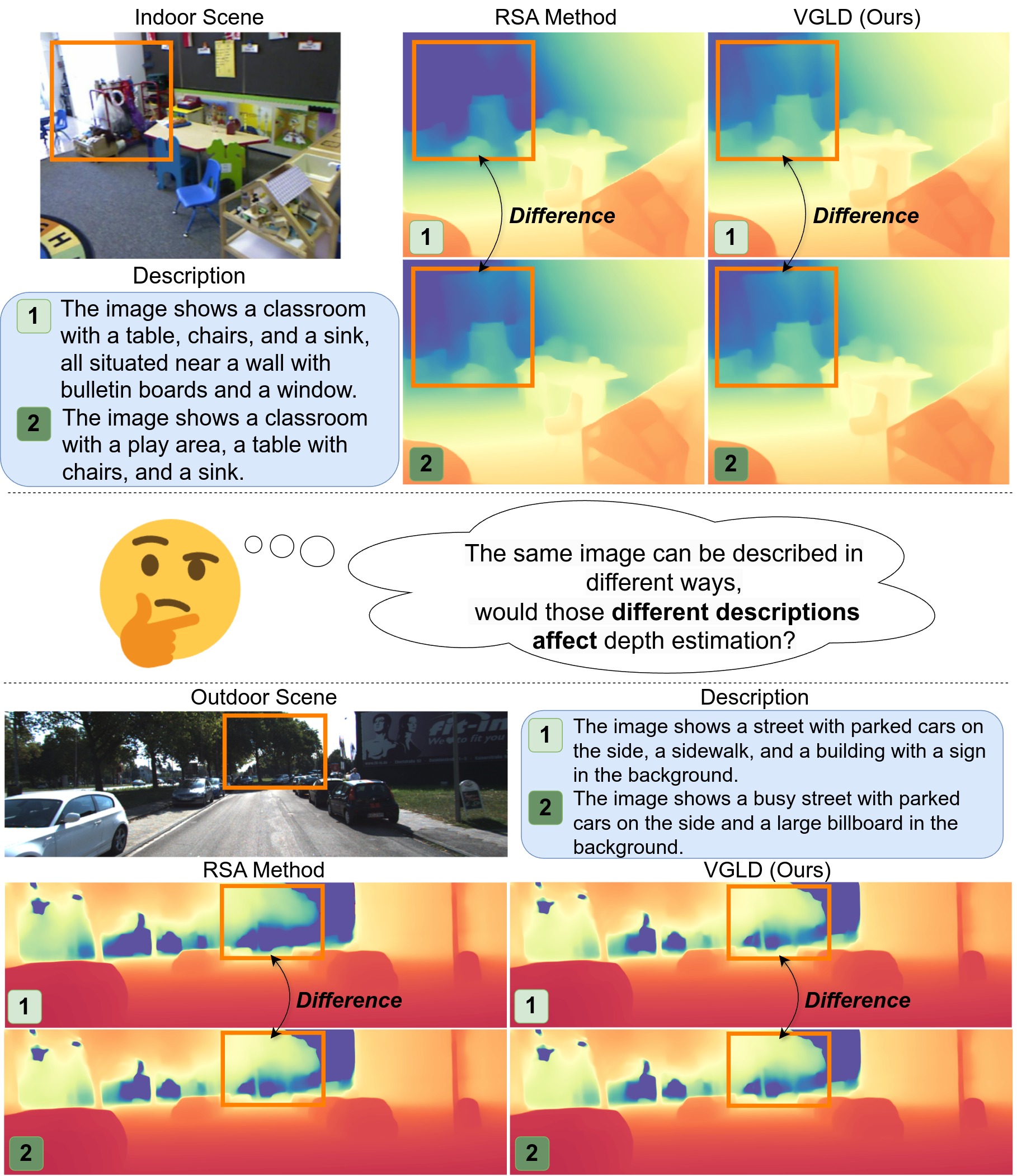} 
\caption{As observed in the figure above, a single image can have multiple different descriptions, and these varying descriptions can significantly affect depth estimation. In particular, the orange bounding boxes in the depth estimation maps highlight this issue, especially for the RSA method, where two semantically similar text descriptions result in substantial differences in depth estimation. In contrast, VGLD(ours) demonstrates relatively stable performance across different descriptions.}
\label{fig1}
\end{figure}

\section{Introduction}

Monocular depth estimation is a fundamental and long-standing task in computer vision, with applications ranging from autonomous driving\cite{schon2021mgnet}, augmented reality\cite{ganj2023mobile} to 3D reconstruction\cite{mescheder2019occupancy}. 
The goal is to predict dense depth maps from single RGB images. However, reconstructing 3D geometry from a single image is an ill-posed problem because perspective projection causes a loss of depth dimension: any point along a projection ray corresponds to the same image coordinate. 
Consequently, the absolute distance from the camera to the scene cannot be directly recovered from a single view. Without camera calibration, additional sensors (e.g., IMU\cite{wofk2023monocular}, LiDAR\cite{lin2024prompting}), or strong priors such as pre-trained depth models, scale ambiguity arises.
While stereo or multi-view images can resolve scale by localizing points in 3D space, modern monocular depth estimation models are often trained on diverse datasets with varying data types and distributions—including single RGB images, video streams, and images with or without calibration parameters. 
These differences exacerbate the challenge of scale ambiguity, especially when deploying models across domains such as indoor and outdoor scenes.

To address scale ambiguity in monocular depth estimation, one line of work trains on multi-domain datasets (e.g., indoor and outdoor) to learn depth from domain-specific distributions\cite{ranftl2020towards, birkl2023midas, yang2024depth, yang2024depthv2}. 
However, dataset biases limit generalization\cite{piccinelli2024unidepth}. An alternative strategy is to leverage complementary cues shared across domains.
Recent approaches explore language as a modality to resolve scale ambiguity without requiring expensive sensors (e.g., LiDAR). 
RSA\cite{zeng2024rsa} pioneers this direction by hypothesizing that textual descriptions can guide scale estimation and demonstrates that scale-less relative depth can be mapped to metric predictions via a language-guided global transformation.

Nevertheless, linguistic inputs are inherently ambiguous—semantically similar captions may produce inconsistent scales (see Figure \ref{fig1}), affecting stability. Still, language is robust to visual challenges like lighting or occlusion.

To reduce linguistic ambiguity, we propose a Visually-Guided Linguistic Disambiguation (VGLD) framework, which enriches textual inputs with semantic features extracted from the corresponding image using a CLIP Image Encoder\cite{radford2021learning}.
Additionally, to handle cross-domain depth variation, we introduce a Domain Router Mechanism (DRM) inspired by ZoeDepth\cite{bhat2023zoedepth}, which routes inputs to domain-specific heads for consistent metric predictions.
To further stabilize training, we formulate depth scale recovery as a scalar regression task and supervise it using pseudo-labels $(k_{lm}, b_{lm})$ obtained via the Levenberg-Marquardt algorithm. This nonlinear optimization technique helps guide the model toward an accurate training trajectory, enhancing robust scale recovery.

Our contributions are as follows:
\begin{itemize}
\item We integrate high-level semantic information from the corresponding image alongside the textual description, thereby stabilizing the output of the scalars parameters;
\item We introduce the Domain Router Mechanism, which aids in solving the cross-domain estimation problem;
\item We leverage the Levenberg-Marquardt algorithm to optimize the training trajectory and guide the model's training process;
\item Extensive experiments demonstrate the effectiveness of our method in both indoor and outdoor scenarios, highlighting its robustness to textual variations and strong zero-shot generalization.
\end{itemize}

\begin{figure*}[!ht]
\centering
\includegraphics[width=\textwidth]{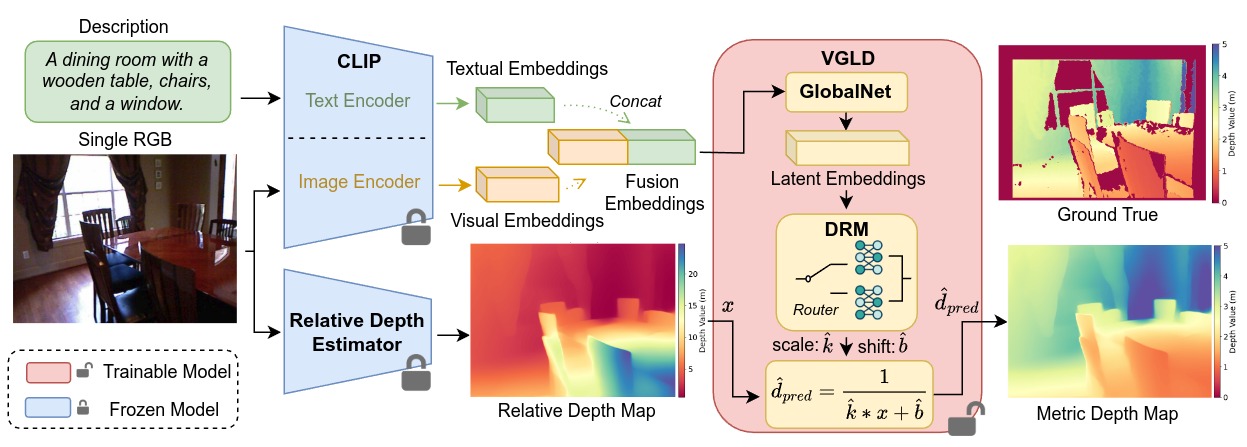}
\caption{Overview. We infer the scale \(\hat{k}\) and shift \(\hat{b}\) from the linguistic description and the corresponding image to transform the relative depth from the depth model into a metric depth (absolute depth in meters) prediction.}  
\label{fig_2}
\end{figure*}

\section{Related Work}
\subsection{Monocular Depth Estimation}
Monocular Depth Estimation (MDE) is a fundamental task in computer vision, with its development generally following two main directions: relative depth estimation and metric depth estimation.
The goal of metric depth estimation is to predict pixel-wise depth values in metric units (e.g., meters), 
and models are typically trained by minimizing the discrepancy between predicted and ground-truth depth maps. 
In contrast, relative depth estimation focuses on inferring the ordinal relationships between pixel pairs, without providing any information about scale or units.
A notable early milestone in this field was Eigen \textit{et al.}\cite{eigen2014depth}, the first to apply Convolutional Neural Networks (CNNs) to MDE. 
More recent methods such as AdaBins \cite{bhat2021adabins}, LocalBins\cite{bhat2022localbins} and Binsformer\cite{li2024binsformer} reformulate the depth regression problem as a classification task through depth discretization. 
Multi-task learning strategies have also been explored: GeoNet\cite{qi2018geonet} integrates surface normal estimation, while AiT\cite{ning2023all} incorporates instance segmentation, both to enhance depth prediction through joint training. 
MiDaS\cite{ranftl2020towards, birkl2023midas} and Diversedepth\cite{yin2020diversedepth} advances relative depth estimation by pretraining on a diverse mixture of datasets, achieving strong generalization across domains. 
In addition, diffusion-based\cite{viola2024marigold, zhang2024betterdepth, song2025depthmaster} methods, such as DDP \cite{ji2023ddp}, Marigold \cite{ke2024repurposing}, 
and GeoWizard \cite{fu2025geowizard}, adapt powerful diffusion priors to the depth estimation task via fine-tuning, enabling significant performance gains.

\subsection{Metric Depth Scale Recovery}
Relative depth estimation models have emerged as strong backbones for many metric depth Scale Recovery tasks, owing to their impressive cross-domain generalization and robustness. 
Building on MiDaS\cite{ranftl2020towards}, DPT\cite{ranftl2021vision} replaces the convolutional backbone with a Vision Transformer and adapts it to metric depth via fine-tuning on scale-annotated datasets. 
ZoeDepth\cite{bhat2023zoedepth} further enhances this pipeline by introducing a powerful decoder with a metric bins module, enabling effective scale recovery through supervised fine-tuning. 
Depth Anything extends ZoeDepth\cite{bhat2023zoedepth} by replacing the MiDaS\cite{ranftl2020towards} encoder with its own architecture, achieving implicit conversion from relative to metric depth.

Other methods like Metric3D\cite{hu2024metric3d, yin2023metric3d}, zeroDepth\cite{guizilini2023towards} and UniDepth\cite{piccinelli2024unidepth} recover scale by leveraging or predicting camera intrinsics, while PromptDA\cite{lin2024prompting} introduces a lightweight LiDAR prompt to guide metric estimation. 
RSA\cite{zeng2024rsa} proposes an alternative paradigm by aligning relative depth with metric scale using textual descriptions, enabling generalization without requiring ground-truth depth at inference. 
However, RSA\cite{zeng2024rsa} is sensitive to linguistic variations, where semantically similar but differently worded inputs may cause inconsistent predictions.
In contrast, VGLD leverages visual semantics to guide linguistic disambiguation, enabling more robust and reliable scale recovery. 
By grounding ambiguous textual inputs in high-level visual context, it mitigates sensitivity to language variation and achieves consistent metric depth estimation across domains.

\subsection{Language Modality for Metric Depth Estimation}

Recent advances in vision-language models\cite{li2022blip, radford2021learning, jia2022visual}, driven by large-scale pretraining, have enabled strong cross-modal representations and inspired new approaches in monocular depth estimation.
DepthCLIP\cite{zhang2022can} first applied CLIP\cite{radford2021learning} to this task by reformulating depth regression as distance classification using natural language descriptions such as \emph{"This object is giant, close...far..."}, enabling zero-shot depth prediction via CLIP's semantic priors.
Subsequent works improved adaptability in various ways: Auty \textit{et al.}\cite{auty2023learning} introduced learnable prompts to replace fixed text tokens; Hu \textit{et al.}\cite{hu2024learning} employed codebooks to address domain shifts; and CLIP2Depth\cite{kim2024clip} proposed mirror embeddings to eliminate reliance on explicit textual input.
Other approaches such as VPD\cite{zhao2023unleashing} , TADP\cite{kondapaneni2024text} , EVP\cite{lavreniuk2023evp} and GeoWizard\cite{fu2025geowizard} extract semantic priors from pretrained text-to-image diffusion models to support depth prediction.

Recently, Wordepth\cite{zeng2024wordepth} modeled language as a variational prior by explicitly encoding object attributes (e.g., size, position) to align relative predictions with metric depth.
RSA\cite{zeng2024rsa} introduced a direct constraint to recover metric scale from text, but suffers from sensitivity to linguistic variation.
In contrast, VGLD combines CLIP-based visual semantics with textual input, offering more stable and robust scale predictions compared to purely language-based methods.

\begin{figure*}[!ht]
\centering
\includegraphics[width=\textwidth]{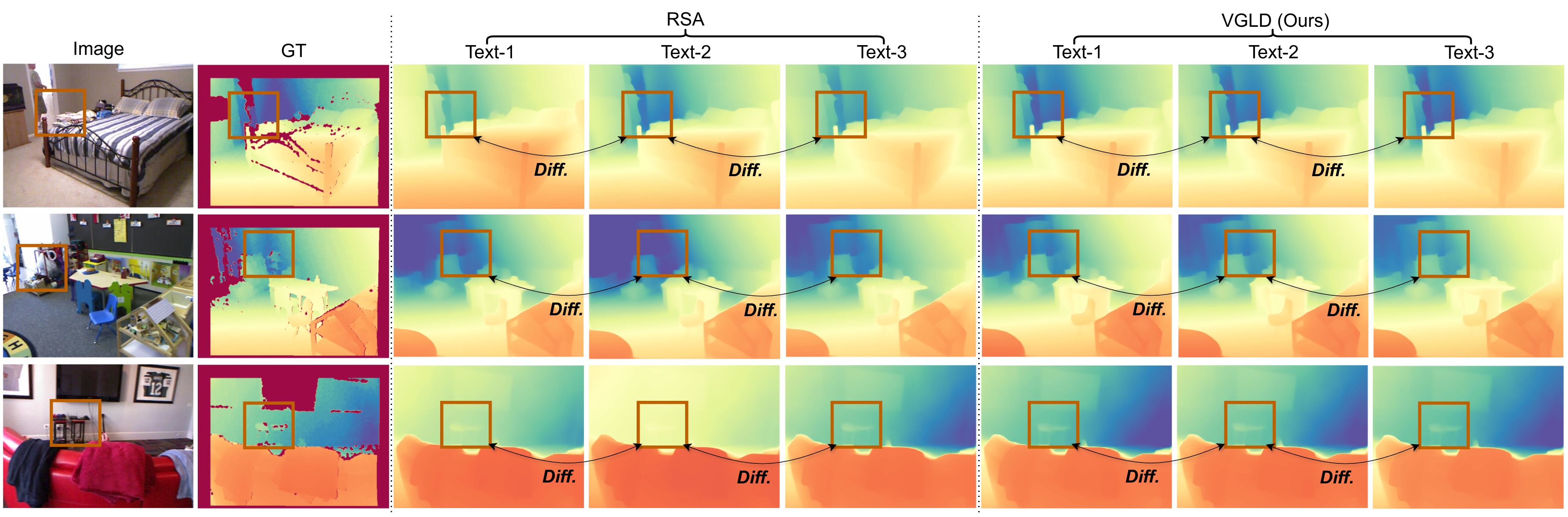}
\caption{Sensitivity to variations in linguistic descriptions on the NYUv2 dataset. 
We focus on the estimation results under three different textual inputs (text1-3). 
As shown in the depth maps, the RSA method exhibits noticeable sensitivity to textual variations, leading to inconsistent predictions—particularly in the regions highlighted by orange boxes. 
In contrast, our proposed VGLD produces more stable and consistent depth estimates across different descriptions.
Warmer colors (red) indicate closer distances, while cooler colors (blue) indicate farther distances.}
\label{fig_diffnyu}
\end{figure*}

\begin{table*}[!h]
\centering
\setlength{\tabcolsep}{2mm}  
{\fontsize{9}{10}\selectfont
\begin{tabular}{l|l|ccc|ccc}
\hline
\multirow{2}{*}{{Models}$\dagger$} & \multirow{2}{*}{{Method}$\ast$} & \multicolumn{3}{c|}{NYUV2} & \multicolumn{3}{c}{KITTI} \\  
& & {Abs Rel $\downarrow$} & {RMSE $\downarrow$} & {D1 $\uparrow$}& {Abs Rel $\downarrow$} & {RMSE $\downarrow$} & {D1 $\uparrow$}\\
\hline
ZoeDepth\cite{bhat2023zoedepth} & \multirow{3}{*}{\centering robust depth estimation $\ddagger$} & 0.077 & 0.277 & 0.953 & 0.054 & 2.281 & 0.971 \\
ZeroDepth\cite{guizilini2023towards} & & 0.074 & 0.269 & 0.954 & 0.053 & 2.087 & 0.968 \\
Metric3Dv2\cite{hu2024metric3d} & & 0.047 & 0.183 & 0.989 & 0.044 & 1.985 & 0.985  \\
\hline

\multirow{10}{*}{MiDas-1\cite{birkl2023midas}}   
& Least Squares & 0.121 & 0.388 & 0.866 & 0.333 & 6.901 & 0.408 \\
& Levenberg Marquardt & 0.056 & 0.218 & 0.969 & 0.091 & 3.373 & 0.925 \\
\cline{2-8}
& RSA-N/K\cite{zeng2024rsa} & 0.171 & 0.569 & 0.731 & 0.163 & 4.082 & 0.798 \\
& RSA-NK\cite{zeng2024rsa} & 0.168 & 0.561 & 0.737 & 0.160 & 4.232 & 0.782 \\
& VGLD-N/K-T (Ours) & 0.158 & 0.529 & 0.758 & 0.133 & 3.755 & 0.854\\
& VGLD-N/K-I (Ours) & 0.121 & \underbar{0.423} & 0.860& \textbf{0.120} & 3.668 & 0.868 \\
& VGLD-N/K-TCI (Ours) & \textbf{0.119} & \textbf{0.414} & \textbf{0.867} & \textbf{0.120} & 3.598 & \underbar{0.871} \\
& VGLD-NK-T (Ours) & 0.159 & 0.526 & 0.751 & 0.130 & 3.744 & 0.844 \\
& VGLD-NK-I (Ours) & 0.123 & 0.426 & 0.855 & \underbar{0.122} & \underbar{3.574} & 0.868  \\
& VGLD-NK-TCI (Ours) & \underbar{0.120} & \textbf{0.414} & \underbar{0.863} & \textbf{0.120} & \textbf{3.559} & \textbf{0.874} \\
\hline

\multirow{8}{*}{MiDas-2\cite{ranftl2020towards}}  
& Least Squares & 0.130 & 0.421 & 0.845 & 0.336 & 6.925 & 0.421 \\
& Levenberg Marquardt & 0.094 & 0.330 & 0.916 & 0.155 & 4.190 & 0.809 \\
\cline{2-8}
& VGLD-N/K-T (Ours) & 0.180 & 0.596 & 0.688 & 0.194 & 5.030 & 0.709 \\
& VGLD-N/K-I (Ours) & \underbar{0.154} & 0.524 & 0.775 & 0.183 & 4.842 & \underbar{0.942} \\
& VGLD-N/K-TCI (Ours) & \textbf{0.151} & \textbf{0.507} & \textbf{0.789} & \textbf{0.178} & \underbar{4.806} & \textbf{0.748} \\
& VGLD-NK-T (Ours) & 0.182 & 0.615 & 0.682 & 0.191 & 4.994 & 0.723 \\
& VGLD-NK-I (Ours) & 0.155 & 0.520 & 0.776 & 0.184 & 4.808 & \underbar{0.740} \\
& VGLD-NK-TCI (Ours) & \textbf{0.151} & \underbar{0.513} & \underbar{0.780} & \underbar{0.180} & \textbf{4.804} & 0.737 \\

\hline
\multirow{8}{*}{DAV2-vits\cite{yang2024depthv2}}  
& Least Squares & 0.122 & 0.392 & 0.866 & 0.330 & 6.737 & 0.423 \\
& Levenberg Marquardt & 0.052 & 0.209 & 0.969 & 0.103 & 3.277 & 0.919 \\
\cline{2-8}
& VGLD-N/K-T (Ours) & 0.163 & 0.546 & 0.713 & {0.166} & {4.189} & {0.756} \\
& VGLD-N/K-I (Ours) & 0.128 & \underbar{0.433} & \underbar{0.830} & {0.154} & {4.219} & {0.756} \\
& VGLD-N/K-TCI (Ours) & \textbf{0.125} & \textbf{0.423} & \textbf{0.842} & \textbf{0.152} & \textbf{3.872} & \textbf{0.779} \\
& VGLD-NK-T (Ours) & 0.161 & 0.539 & 0.714 & {0.164} & {4.287} & {0.752} \\
& VGLD-NK-I (Ours) & \underbar{0.127} & 0.436 & 0.835 & {0.160} & {4.031} & {0.761} \\
& VGLD-NK-TCI (Ours) & \underbar{0.127} & 0.434 & 0.835 & \underbar{0.153} & \underbar{3.980} & \underbar{0.772} \\
\hline

\multirow{10}{*}{DAV1-vits\cite{yang2024depth}}  
& Least Squares & 0.121 & 0.397 & 0.863 & 0.331 & 6.772 & 0.423 \\
& Levenberg Marquardt & 0.057 & 0.230 & 0.967 & 0.112 & 3.375 & 0.897 \\
\cline{2-8}
& RSA-N/K\cite{zeng2024rsa} & 0.147 & 0.484 & 0.775 & {0.160} & {4.437} & {0.780} \\
& RSA-NK\cite{zeng2024rsa} & 0.148 & 0.498 & 0.776 & {0.158} & {4.457} & {0.786} \\
& VGLD-N/K-T (Ours) & 0.145 & 0.496 & 0.792 & {0.151} & {4.354} & {0.773} \\
& VGLD-N/K-I (Ours) & 0.115 & 0.405 & 0.872 & {0.144} & \underbar{4.074} & {0.790} \\
& VGLD-N/K-TCI (Ours) & \textbf{0.112} & \textbf{0.390} & \textbf{0.887} & \underbar{0.140} & {4.081} & {0.807} \\
& VGLD-NK-T (Ours) & 0.142 & 0.483 & 0.787 & {0.148} & {4.293} & {0.781} \\
& VGLD-NK-I (Ours) & \underbar{0.114} & 0.404 & 0.880 & {0.142} & {4.151} & \underbar{0.814} \\
& VGLD-NK-TCI (Ours) & \textbf{0.112} & \underbar{0.392} & \underbar{0.883} & \textbf{0.136} & \textbf{4.008} & \textbf{0.816} \\
\bottomrule
\end{tabular}}
\caption{Quantitative Depth Comparison on the NYUV2 and KITTI Dataset. 
$\dagger$ In the Model column, MiDas-1 denotes Midas-V3.1-dpt\_swin2\_large\_384, MiDas-2 denotes Midas-V3.0-dpt\_large\_384, DAV2-vits denotes Depth-Anything-V2-Small, and DAV1-vits denotes Depth-Anything-V1-Small. 
$\ddagger$ denotes the results of certain state-of-the-art (SOTA) absolute scale estimation models. 
$\ast$ In the Method column, “N” and “K” indicate models trained on the NYUv2 and KITTI datasets, respectively. For example, VGLD-N/K-TCI refers to VGLD-N-TCI when evaluated on NYUv2, and VGLD-K-TCI when evaluated on KITTI.
Best results are in \textbf{bold}, second best are \underline{underlined}.
}
\label{tab:1}
\end{table*}


\section{Method}
\subsection{Preliminaries}
The objective of monocular depth estimation is to predict continuous per-pixel depth values from a single RGB image\cite{eigen2014depth}. 
We consider a dataset \( \mathcal{D} = \{(I^{(n)}, t^{(n)}, d^{(n)}_{gt}, dm^{(n)}_{gt})\}_{n=1}^N \) consisting of \( N \) samples, 
where each sample includes an RGB image \( I \in \mathbb{R}^{3 \times H \times W} \), a corresponding linguistic description \( t \), 
a ground-truth metric depth map \( d_{gt} \in \mathbb{R}^{H \times W} \) and a ground-truth domain labels \(dm_{gt} \in \{0, 1\}\) which represent \textit{indoor} or \textit{outdoor} scene.
We build upon a pretrained monocular relative depth estimation model \( h_\theta \), 
which serves as the foundation for our metric depth scale recovery framework. Given an RGB image, 
the model predicts an inverse relative depth map \( x \in \mathbb{R}^{H \times W} \), 
which lacks absolute scale information. To recover metric-scale depth from this scaleless prediction, 
we apply a global linear transformation informed by both the linguistic description and high-level visual semantics of the image. 
Specifically, similar to RSA\cite{zeng2024rsa}, we predict a pair of scalars \( (\hat{k}, \hat{b}) \in \mathbb{R}^2 \) that represent the scale and shift parameters of the transformation. 
The final metric depth prediction is then computed as: 
\begin{equation}
\label{depth recovery}
\hat{d}_{pred} = \frac{1}{\hat{k} \cdot x + \hat{b}} \text{  ,where  } \hat{d}_{pred} \in \mathbb{R}^{H \times W}
\end{equation}

\subsection{VGLD}
To model the relationship between the linear transformation parameters and the semantic content of both the image and its linguistic description, 
we leverage the CLIP model as a feature extractor. Benefiting from large-scale contrastive pretraining\cite{radford2021learning}, 
CLIP provides a shared latent space that is well-suited for aligning object-centric visual and linguistic representations. 
Given an input sample \( \{I, t\} \), we first extract visual and text embeddings using the CLIP image encoder and CLIP text encoder, respectively. 
The resulting embeddings are concatenated to form a fused representation, which is subsequently passed through a lightweight encoder network, 
GlobalNet—a three-layer MLP—to produce a compact 256-dimensional latent embedding used for downstream scale parameter regression.

Following ZoeDepth\cite{bhat2023zoedepth}, we employ a lightweight MLP-based classifier, referred to as the Domain Routing Mechanism (\textbf{DRM}), 
to predict the domain of the input image based on its latent embedding. We consider two domains: indoor and outdoor. 
The predicted domain is then used to route the latent embedding to the corresponding domain-specific scalars prediction head.

\subsection{Loss Function}
As illustrated in Figure \ref{fig_2}, the VGLD model freezes the weights of both the CLIP backbone and the relative depth estimator 
during training, and updates only the parameters of the GlobalNet and DRM modules. These modules are jointly optimized under a unified loss function. 
Since VGLD focuses on predicting a pair of global scalars rather than pixel-wise metric depth values, we do not adopt the Scale-Invariant Logarithmic Loss, 
which is more suitable for dense depth estimation tasks. Instead, following RSA\cite{zeng2024rsa}, we adopt the L1 loss, which provides a more direct and interpretable supervision signal for scalars regression. 
The $\mathcal{L}_{metric}$ is formulated as:
\begin{equation}
\label{metric_loss}
\mathcal{L}_{metric} = \frac{1}{M} \sum_{(i, j) \in \Omega} m(i,j) \times |\hat{d}_{pred}(i, j) - d_{gt}(i, j)|,
\end{equation}
where \( \hat{d}_{pred} \) denotes the predicted metric depth, \( (i, j) \in \Omega \) represents the image coordinates, \(m(\cdot ) \in \{0,1\}\)  denotes the binary mask map and \(M\)
represents the number of pixels with valid ground truth values.

To ensure correct routing to the domain-specific scalars prediction head, We introduce a domain classification loss, denoted as \( \mathcal{L}_{\text{dm}} \), implemented using the cross-entropy loss:
\begin{equation}
\label{domain_loss}
\mathcal{L}_{domain} = CrossEntropy(\hat{dm}_{pred}, dm_{gt})
\end{equation}
where \( \hat{dm}_{pred} \in \{0, 1\}\) is the predicted domain label, and \( dm_{gt} \in \{0, 1\} \) is the corresponding ground-truth domain.

To guide the model towards the optimal solution, we employ an MSE loss to provide LM loss (scalars supervision) for the modules:
\begin{equation}
\label{scalars_loss}
\mathcal{L}_{lm} = 10 \times (\hat{k} - k_{\text{lm}})^2 + (\hat{b} - b_{\text{lm}})^2
\end{equation}
where \( (\hat{k}, \hat{b}) \) are the predicted LM scalars from VGLD, and \( (k_{\text{lm}}, b_{\text{lm}}) \) are the corresponding pseudo-labels provided by the Levenberg-Marquardt algorithm.
We assign a higher weight (10x) to the scale term \(k_{\text{lm}}\) because empirical observations show that the model is more sensitive to errors in scale prediction than in shift. This design choice helps stabilize training and ensures more accurate depth scaling.

The total loss is defined as follows:
\begin{equation}
\label{total_loss}
\mathcal{L}_{total} = \mathcal{L}_{metric} + \alpha \times \mathcal{L}_{domain} + \beta \times \mathcal{L}_{lm}
\end{equation}
In our experiments, we set \(\alpha\) and \(\beta\) to 0.1, as is customary.

\begin{figure*}[!ht]
\centering
\includegraphics[width=\textwidth]{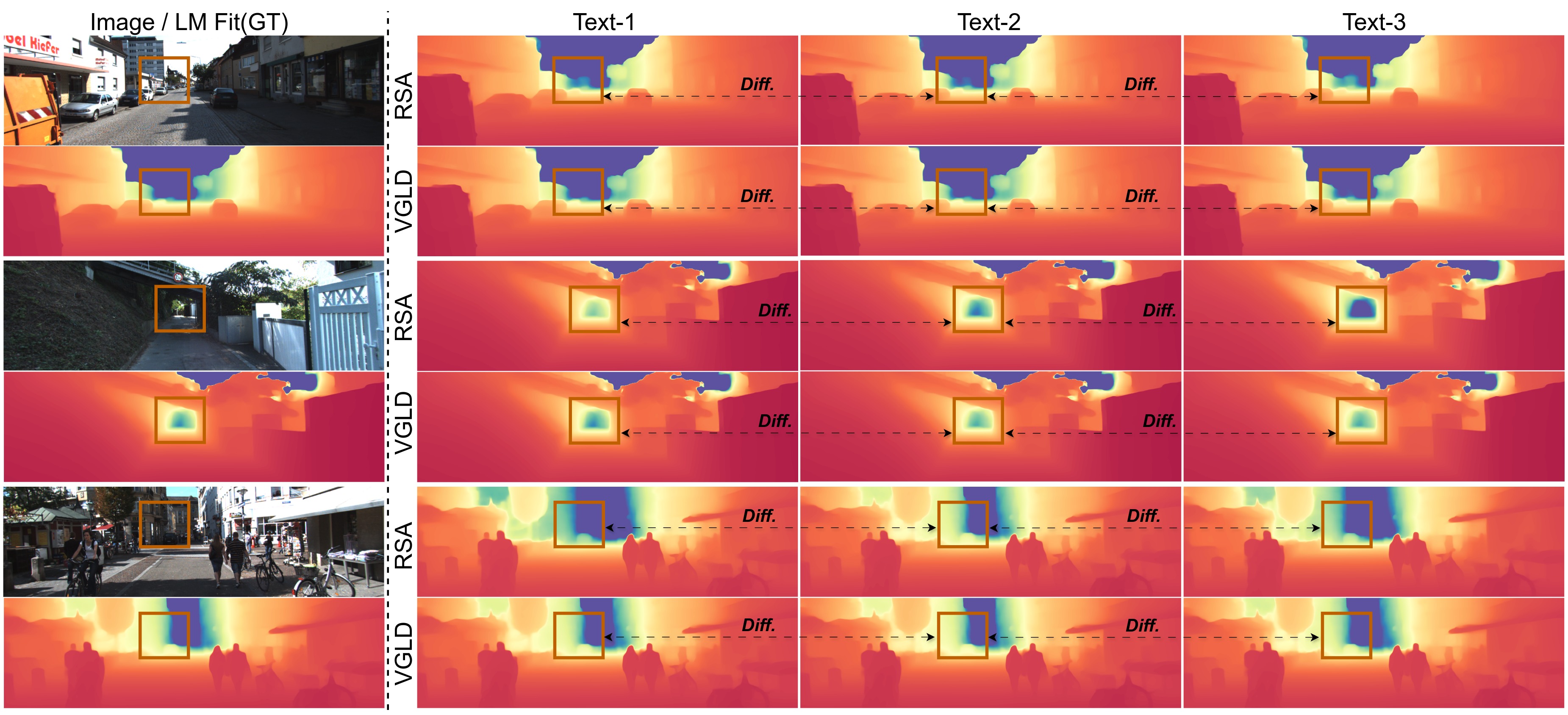}
\caption{Sensitivity to variations in linguistic descriptions on the KITTI dataset. 
Similar to Figure \ref{fig_diffnyu}, we focus on the differences within the orange boxes across the three textual inputs. Note that we use LM fitting results instead of the ground-truth depth map for visualization, as the KITTI ground-truth data is too sparse to yield meaningful visual comparisons.
Warmer colors (red) indicate closer distances, while cooler colors (blue) indicate farther distances.}
\label{fig_diffkitti}
\end{figure*}

\section{Experiments}
\subsection{Experimental Settings}
\textbf{Dataset.}
We primarily train on two datasets: NYUv2\cite{silberman2012indoor} and KITTI\cite{geiger2012we}, representing indoor and outdoor scenes, 
respectively. NYUv2 contains images with a resolution of 480x640, with depth values ranging from 0 to 10 meters. 
In accordance with the official dataset split\cite{lee2019big}, we use 24,231 image-depth pairs for training and 654 image-depth pairs for testing. 
KITTI is an outdoor dataset collected from equipment mounted on a moving vehicle, with depth values ranging from 0 to 80 meters. 
Following KBCrop\cite{uhrig2017sparsity}, all RGB images and depth maps are cropped to a resolution of 1216x352. 
We adopt the Eigen split\cite{eigen2014depth}, which includes 23,158 training images and 652 test images, to train and evaluate our method. 
Additionally, we report zero-shot generalization results on SUNRGBD\cite{song2015sun}, which includes 5,050 test images, DIML Indoor\cite{cho2021diml}, 
which contains 503 validation images and DDAD\cite{guizilini20203d}, which contains 3950 validation images.

\noindent \textbf{Relative Depth Models.} 
We use MiDaS 3.1\cite{birkl2023midas} with the \textit{dpt\_swin2\_large\_384} model (213M parameters),  MiDaS 3.0\cite{ranftl2020towards} with the \textit{dpt\_large\_384} model (123M parameters), 
DepthAnything\cite{yang2024depth} with \textit{DepthAnything-Small} model (24.8M parameters), and DepthAnything v2\cite{yang2024depthv2}  with \textit{DepthAnything-V2-Small} model (24.8M parameters). 

\noindent \textbf{The Proposed Models.} For clarity, we denote the proposed models as VGLD-\{dataset\}-\{method\}. The \{dataset\} refers to the training datasets, 
which include "N" for NYUv2, "K" for KITTI, and "NK" for both NYUv2 and KITTI. The \{method\} refers to the type of embeddings used: "T" for text embeddings only, 
"I" for visual embeddings only, and "TCI" for both text and visual embeddings (i.e., Fusion Embeddings, as shown in Figure \ref{fig_2}).

\noindent \textbf{Evaluation details.}
We evaluate performance using several metrics, including mean absolute relative error (Abs Rel), squared relative error (sq\_rel), 
root mean square error (RMSE), root mean square error in log space (\(\text{RMSE}_{\text{log}}\)), absolute error in log space (\(\text{log}_{\text{10}}\)) and threshold accuracy (\(\delta_i\)).

\subsection{Experimental Results}
\textbf{Quantitative results.}
We present the results on the NYUv2 and KITTI datasets in Table \ref{tab:1}. (More detailed quantitative results are provided in Table \ref{tab:Appendix_nyu} and Table \ref{tab:Appendix_kitti} in the Supplementary Material.). Our approach consistently outperforms RSA\cite{zeng2024rsa} across all evaluation metrics 
and achieves performance comparable to scale recovery using ground-truth depths, as indicated in the Least Squares and Levenberg-Marquardt sections of the quantitative tables.
The quantitative results show that models trained on a single dataset (VGLD-N or VGLD-K) perform slightly better within their respective domains compared to the unified model VGLD-NK. 
For example, VGLD-N/K-TCI with DAV2-ViTS as the RDE model achieves the best performance across all three evaluation metrics reported in the Table \ref{tab:1}.
Thanks to the precise routing capability of the DRM module, the performance gap between the single-dataset and unified models remains marginal, highlighting the strong cross-domain generalization ability of the unified VGLD-NK model.
For example, based on DAV2-ViTS, the VGLD-N-TCI model achieves an AbsRel of 0.125 on NYUv2, and the VGLD-K-TCI model achieves 0.152 on KITTI. The unified VGLD-NK-TCI model obtains AbsRel scores of 0.127 and 0.153 on NYUv2 and KITTI, respectively, representing decreases of less than 1.58\% and 0.65\%.

Furthermore, models utilizing visual embeddings (VGLD-XX-I) consistently outperform those relying solely on textual embeddings (VGLD-XX-T), validating the effectiveness of visual cues for scale prediction over purely linguistic prompts. 
For example, based on DAV1-ViTS, the VGLD-NK-T model achieves AbsRel scores of 0.142 and 0.148 on NYUv2 and KITTI, respectively. In comparison, VGLD-NK-I achieves AbsRel scores of 0.114 and 0.142 on the same datasets, corresponding to improvements of 24.5\% and 4.2\%, respectively.
Building on this, we combine both visual and textual embeddings (VGLD-XX-TCI), allowing visual features to guide the semantic alignment of textual inputs. This integration yields modest but meaningful improvements, thereby effectively addressing the challenge of visually grounded linguistic disambiguation.

Notably, the improvement of VGLD-XX-TCI over VGLD-XX-T is less pronounced on KITTI compared to NYUv2. We attribute this to the lower variance in outdoor scene descriptions in KITTI, 
whereas indoor scenes in NYUv2 exhibit much greater diversity—such as bathrooms, kitchens, classrooms... This higher variability in textual descriptions benefits the model by providing richer cues for more accurate estimation of scene-specific scaling parameters.

For completeness, the Supplementary Material presents more extensive quantitative results and qualitative comparisons, including those from the zero-shot evaluation setting.

\noindent \textbf{Sensitivity to Variations in Linguistic Descriptions.}
A single image can be described using multiple textual expressions. To investigate how linguistic variation affects metric depth scale recovery, we evaluate the influence of different textual inputs on VGLD's performance. 
Figures \ref{fig_diffnyu} and \ref{fig_diffkitti} present qualitative comparisons on NYU and KITTI under three distinct text prompts.
We observe that while the RSA method—relying solely on textual descriptions—is highly sensitive to phrasing, VGLD demonstrates significantly greater robustness, consistently producing stable predictions for both scale and shift. 
This is most evident in the third image of Figure \ref{fig_diffnyu}: RSA accurately recovers the depth when paired with Text-3 (whose prediction closely matches the ground truth), but exhibits substantial errors with Text-1 and Text-2. In contrast, VGLD achieves stable and accurate scale recovery across all three descriptions (Text-1 to Text-3).
Moreover, VGLD often outperforms RSA across evaluation metrics, further highlighting its ability to provide reliable scalar estimations.
The corresponding quantitative results are provided in the Supplementary Material, along with the three textual descriptions used for each image.

\subsection{Ablation Study}
\noindent \textbf{Effect of the DRM.}
As shown in Table \ref{tab:rde_drm_results}, we conduct ablation studies on the Domain Router Mechanism (DRM). 
The results demonstrate that incorporating the DRM consistently improves the overall performance of VGLD across all four backbone models and significantly enhances its cross-domain generalization capability.
The ablation studies are conducted based on the VGLD-XX-TCI model.
\begin{table}[H]
\centering
\setlength{\tabcolsep}{0mm}  
{\fontsize{9}{10}\selectfont
\begin{tabular}{|c|c|ccc|ccc|}
\hline
\multirow{2}{*}{Models} & \multirow{2}{*}{Method} & \multicolumn{3}{c|}{NYU} & \multicolumn{3}{c|}{KITTI} \\
\cline{3-8}
  & & AbsRel$\downarrow$ & RMSE$\downarrow$ & D1$\uparrow$ & AbsRel$\downarrow$ & RMSE$\downarrow$ & D1$\uparrow$ \\
\hline
\multirow{2}{*}{MiDas-1} & w/o DRM & {0.128} & {0.415} & {0.757} & {0.136} & {3.636} & {0.855} \\
  & with DRM   & \textbf{0.120} & \textbf{0.414} & \textbf{0.863} & \textbf{0.121} & \textbf{3.559} & \textbf{0.874} \\
\hline
\multirow{2}{*}{MiDas-2} & w/o DRM & {0.158} & {0.513} & {0.740} & {0.198} & {4.987} & {0.728} \\
  & with DRM   & \textbf{0.151} & \textbf{0.513} & \textbf{0.780} & \textbf{0.185} & \textbf{4.804} & \textbf{0.737} \\
\hline
\multirow{2}{*}{DAV2-vits} & w/o DRM & {0.135} & {0.459} & {0.798} & {0.163} & {4.060} & {0.767} \\
  & with DRM   & \textbf{0.127} & \textbf{0.434} & \textbf{0.835} & \textbf{0.153} & \textbf{3.980} & \textbf{0.772} \\
\hline
\multirow{2}{*}{DAV1-vits} & w/o DRM & {0.122} & {0.437} & {0.847} & {0.145} & {4.327} & {0.748} \\
  & with DRM   & \textbf{0.112} & \textbf{0.392} & \textbf{0.883} & \textbf{0.136} & \textbf{4.008} & \textbf{0.816} \\
\hline
\end{tabular}
}
\caption{Performance comparison on NYU and KITTI datasets with and w/o(without) DRM. Best results are in \textbf{bold}.}
\label{tab:rde_drm_results}
\end{table}

\noindent \textbf{Effect of the LM loss.}
To investigate the effect of different weights of LM loss $\mathcal{L}_{lm}$ on model training, we vary the value of $\beta$ in equation \ref{total_loss} and train the VGLD-NK-TCI model based on the DAV1-vits RDE backbone. 
Evaluation on both the NYUv2 and KITTI datasets shown in Table \ref{tab:alpha_ablation} that the model achieves the best performance when $\beta = 0.1$.
Compared to completely removing the $\beta$ term ($\beta = 0$), the model achieves a 2.7\% improvement in AbsRel on NYUv2 and a significantly larger gain of approximately 20.5\% on KITTI. 
This demonstrates the effectiveness of the $\mathcal{L}_{\text{lm}}$ constraint, particularly in more open outdoor environments, where stronger guidance is needed to stabilize the training trajectory.

\begin{table}[H]
\centering
\setlength{\tabcolsep}{1.4mm}  
{\fontsize{9}{10}\selectfont
\begin{tabular}{|c|ccc|ccc|}
\hline
\multirow{2}{*}{$\beta$} & \multicolumn{3}{c|}{NYU} & \multicolumn{3}{c|}{KITTI} \\
\cline{2-7}
& Abs Rel$\downarrow$ & RMSE$\downarrow$ & D1$\uparrow$ & Abs Rel$\downarrow$ & RMSE$\downarrow$ & D1$\uparrow$ \\
\hline
{0}     & {0.115} & {0.403} & {0.874} & {0.164} & {4.856} & {0.781} \\
{0.001} & {0.116} & {0.413} & {0.869} & {0.161} & {4.204} & {0.779} \\
{0.01}  & \underline{0.113} & {0.399} & \underline{0.879} & \underline{0.146} & \underline{4.010} & \underline{0.791} \\
{0.1}   & \textbf{0.112} & \textbf{0.392} & \textbf{0.883} & \textbf{0.136} & \textbf{4.008} & \textbf{0.816} \\
{1}     & {0.115} & \underline{0.397} & {0.868} & {0.162} & {4.701} & {0.778} \\
\hline
\end{tabular}
}
\caption{Ablation on LM loss for NYUv2 and KITTI datasets. Best results are in \textbf{bold}, second best are \underline{underlined}.}
\label{tab:alpha_ablation}
\end{table}
  
\noindent \textbf{Computational Complexity.}
As shown in Table \ref{tab:vgld_params}, we present a comparison of model parameters and inference times between VGLD and RSA to quantify the computational resources required. 
All evaluations were conducted on a single NVIDIA RTX 3090 (24GB). This experiment is conducted using the DAV1-vits RDE backbone. The results indicate that the scalar predictor in VGLD is more lightweight and efficient compared to that of RSA.
However, VGLD additionally incorporates a CLIP image encoder, which introduces an extra 14ms of inference time compared to RSA. 
Despite this overhead, VGLD offers a favorable trade-off: it achieves a 32.1\%(Ref. to Tabel \ref{tab:1}) improvement in Abs Rel on NYUv2 with an inference time of just 14.08ms increases and a modest parameters, 
making it a practical and efficient choice.

\begin{table}[!ht]
\centering
\setlength{\tabcolsep}{0.8mm} 
{\fontsize{9}{10}\selectfont
\begin{tabular}{|c|cc|cc|}
\hline
\multirow{2}{*}{{Components}} & \multicolumn{2}{c|}{{RSA}} & \multicolumn{2}{c|}{{VGLD} (ours)} \\
\cline{2-5}
& {Params\#} & {Inf. Times} & {Params\#} & {Inf. Times} \\
\hline
DAV1-vits            & 24.78M & 9.62ms  & 24.78M & 9.62ms \\
CLIP Text Encoder    & 63.43M & 13.61ms & 63.43M & 13.61ms \\
CLIP Image Encoder   & -     & -     & 86.19M & 14.90ms \\
Scalars Predictor    & 1.49M  & 1.76ms  & 1.18M  & 0.94ms \\
\hline
\textbf{Total}       & \textbf{89.7M}  & \textbf{24.99ms} & \textbf{175.58M} & \textbf{39.07ms} \\
Increase / M (ms)    & {-}     & {-}     & {85.88M $\uparrow$}  & {14.08ms $\uparrow$} \\
\hline
\end{tabular}
}
\caption{Computational Complexity Analysis.  
As shown in the table, the increase in model parameters(Params\#) and inference times(Inf. Times) of VGLD compared to the RSA model primarily stems from the additional CLIP Image Encoder component.}
\label{tab:vgld_params}
\end{table}


\section{Conclusion}
We presented VGLD, a novel framework for monocular depth scale recovery that performs Visually-Guided Linguistic Disambiguation. 
VGLD leverages high-level visual semantics to resolve inconsistencies in textual inputs, enabling stable and accurate scale prediction across diverse linguistic descriptions. 
By jointly encoding image and text via CLIP and predicting global transformation parameters with an MLP, VGLD transforms relative depth maps into metric estimates in a robust and consistent manner. 
Extensive evaluations on both indoor and outdoor benchmarks show that VGLD significantly reduces estimation variance under different captions and generalizes well across domains. 
Empowered by a Domain Router Mechanism, VGLD further supports universal deployment across scene types. Compared to sensor-based methods, VGLD offers a lightweight and effective alternative for reliable scale alignment.

\section{Limitations and future work.}
Although linguistic-based scale recovery under visually-guided methods is highly robust, VGLD is still influenced by language modality. 
For different descriptions of the same image, the VGLD model may output inconsistent results (albeit with small error margins), especially when incorrect descriptions are used (e.g., describing an indoor scene as \textit{"a photo of a narrow street."}). 
To address this issue, one feasible approach could be to further match the similarity between the language and image modalities, effectively excluding erroneous image descriptions. 
Future work could expand the image modality-assisted features of VGLD to enable more robust and fine-grained scale estimation, as well as enhance the model's ability to handle malicious attacks in text descriptions.

\bibliography{references}

\newpage  
\appendix
\section{Supplementary Material}
\subsection{Evaluation Metrics}
We evaluate our approach using the standard five error metrics and three accuracy metrics commonly adopted in prior works\cite{shao2023urcdc}. 
Specifically, the error metrics include absolute mean relative error (Abs Rel), square relative error (sq\_rel), log error($\log_{10}$),  
root mean squared error (RMSE), and its logarithmic variant (\(\text{RMSE}_{\text{log}}\)). 
The accuracy metrics are based on the percentage of inlier pixels (\(\delta\)) within three thresholds: \(\delta_1 < 1.25\), \(\delta_2 < 1.25^2\), and \(\delta_3 < 1.25^3\).  
\begin{itemize}
\item $\text{Abs Rel}$: $\frac{1}{M} \sum_{(i, j) \in \Omega}|\hat{d}_{pred}(i,j) - d_{gt}(i,j)|/d_{gt}(i,j) $
\item $\text{sq\_rel}$: $\frac{1}{M} \sum_{(i, j) \in \Omega}[(\hat{d}_{pred}(i,j) - d_{gt}(i,j)) / d_{gt}(i,j)]^2 $
\item $\text{RMSE}$: $\sqrt{\frac{1}{M} \sum_{(i, j) \in \Omega}(\hat{d}_{pred}(i,j) - d_{gt}(i,j))^2}$
\item $\text{RMSE}_\text{log}$: $\sqrt{\frac{1}{M} \sum_{(i, j) \in \Omega}(\log \hat{d}_{pred}(i,j) - \log d_{gt}(i,j))^2}$
\item $\text{log}_{10}$: $\frac{1}{M} \sum_{(i, j) \in \Omega} | \log_{10}(\hat{d}_{pred}(i,j)) - \log_{10}(d_{gt}(i,j)) |  $ 
\item $ \text{D} < thr$: $(max(\frac{\hat{d}_{pred}}{d_{gt}}, \frac{d_{gt}}{\hat{d}_{pred}}))$ , $thr = 1.25, 1.25^2, 1.25^3 $
\end{itemize}

\subsection{Training details}
The proposed VGLD is implemented in PyTorch2.0.1+CUDA11.8. We use the Adam optimizer with parameters \((\beta_1, \beta_2, \text{wd}) = (0.9, 0.999, 0.001)\) 
and a learning rate of \(3 \times 10^{-4}\). All models are trained for 24 epochs on a single NVIDIA RTX 3090 GPU with 24GB of memory, , running in Ubuntu 22.04. 
The batch size is set to 6, and the total training time for each model is approximately 19 to 22 hours.

\subsection{Qualitative comparisons}
\begin{figure*}[h]
\centering
\includegraphics[width=0.93\textwidth]{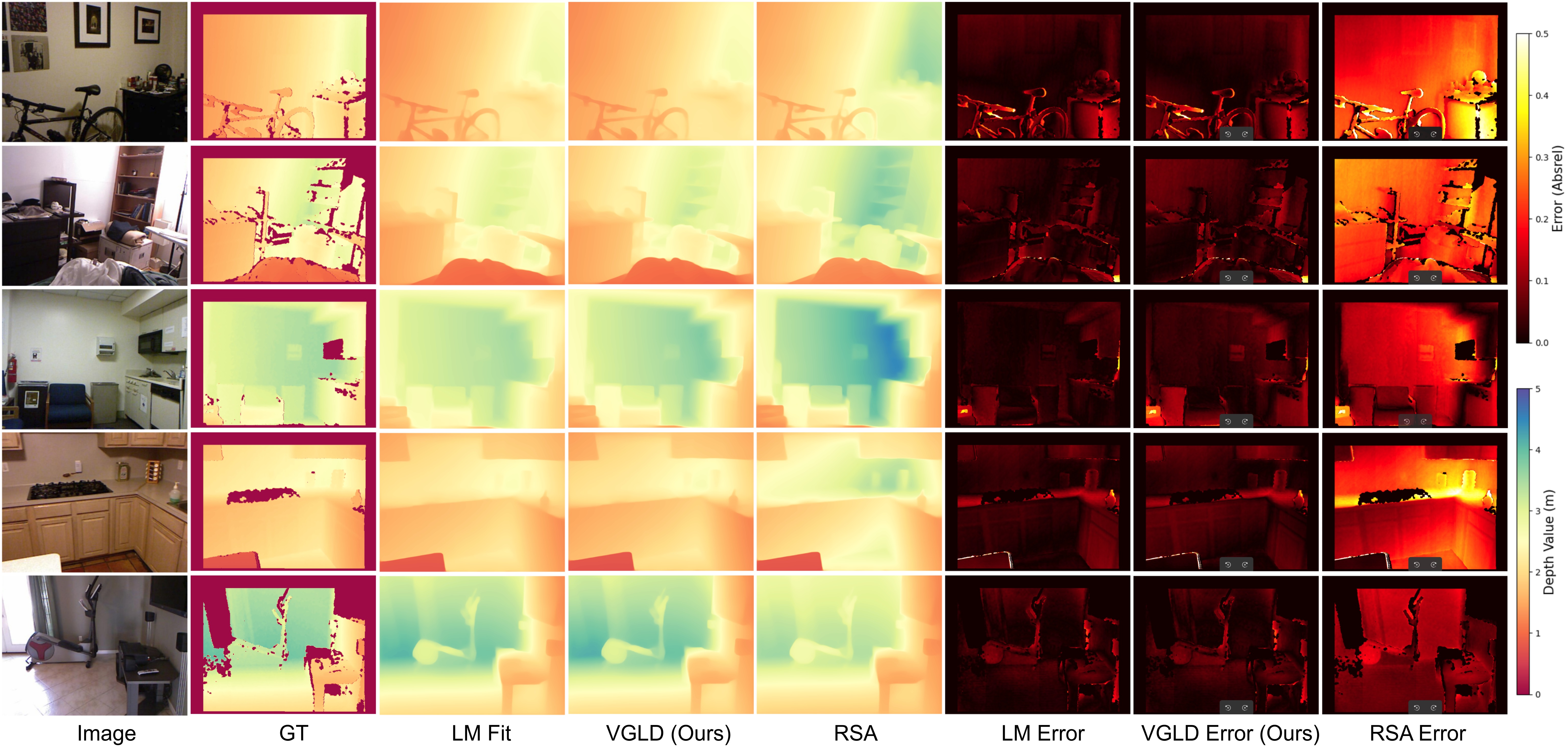}
\caption{Visualization of depth estimation on the NYUv2 dataset. The LM Fit represents the result obtained using the Levenberg-Marquardt algorithm. Note: Zeros in the ground truth indicate the absence of valid depth values (represented in black or deep red).}
\label{fig_vgld_nyu}
\end{figure*}

\begin{figure*}[h]
\centering
\includegraphics[width=0.92\textwidth]{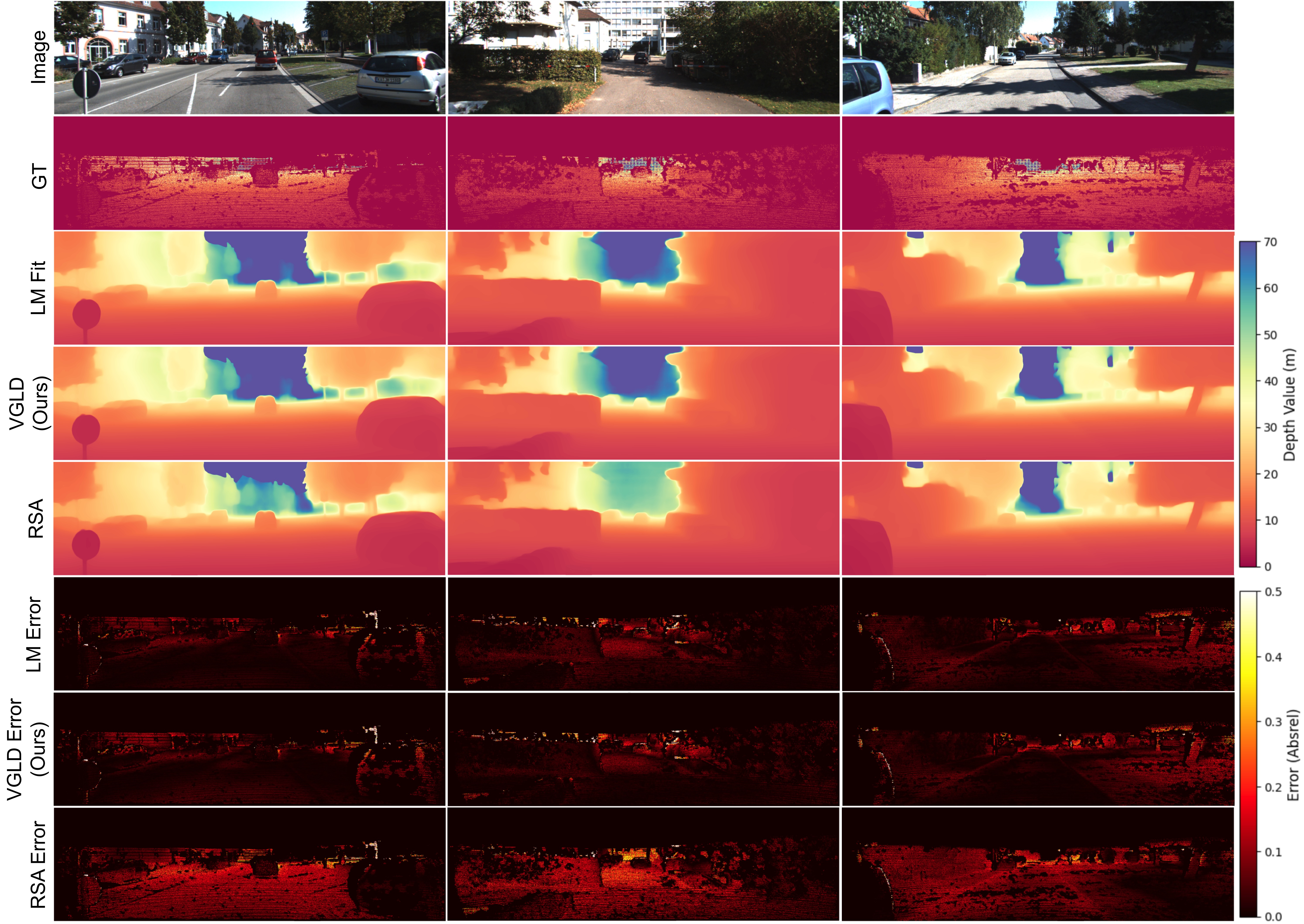}
\caption{Visualization of depth estimation on the KITTI dataset. The LM Fit represents the result obtained using the Levenberg-Marquardt algorithm. Note: Zeros in the ground truth indicate the absence of valid depth values (represented in black or deep red).}
\label{fig_vgld_kitti}
\end{figure*}

We present comparison examples of VGLD and baseline methods on the NYUv2 and KITTI datasets in Figure \ref{fig_vgld_nyu} and Figure \ref{fig_vgld_kitti}, respectively. 
The error maps display the absolute relative error, where the overall brightness of the error maps clearly indicates the performance of our method. 
Notably, our approach achieves performance very close to that of the Levenberg-Marquardt fitting (LM Fit) across different scenes, demonstrating robust metric depth scale recovery. 
In contrast to the fixed scale and shift estimates produced by RSA, VGLD significantly improves the accuracy of depth predictions, with darker error maps indicating reduced error.
Note: All qualitative comparison results in the VGLD section are inferred from the VGLD-NK-TCI method, where the RDE model used is DAV1-vits.

\subsection{Quantitative Results on Sensitivity to Linguistic Description Variations}
As shown in Table \ref{tab:diff nyu} and Table \ref{tab:diff kitti}, We quantitatively evaluated the inference results and sensitivity of the VGLD model to variations in linguistic descriptions. 
For both indoor and outdoor datasets, three images were used, with each image paired with three distinct textual descriptions. 
The corresponding visualization figures are provided in Figure \ref{fig_diffnyu} and Figure \ref{fig_diffkitti}(within the main text)..
And the specific textual descriptions are provided in Table \ref{tab:nyu text} and Table \ref{tab:kitti text}.

From the tables, it is evident that the VGLD model demonstrates greater robustness when processing three different textual descriptions, while the RSA model exhibits larger errors. 
Moreover, under identical textual descriptions, VGLD consistently outperforms RSA.

\subsection{Zero-shot Generalization}
\begin{figure*}[t]
\centering
\includegraphics[width=1\textwidth]{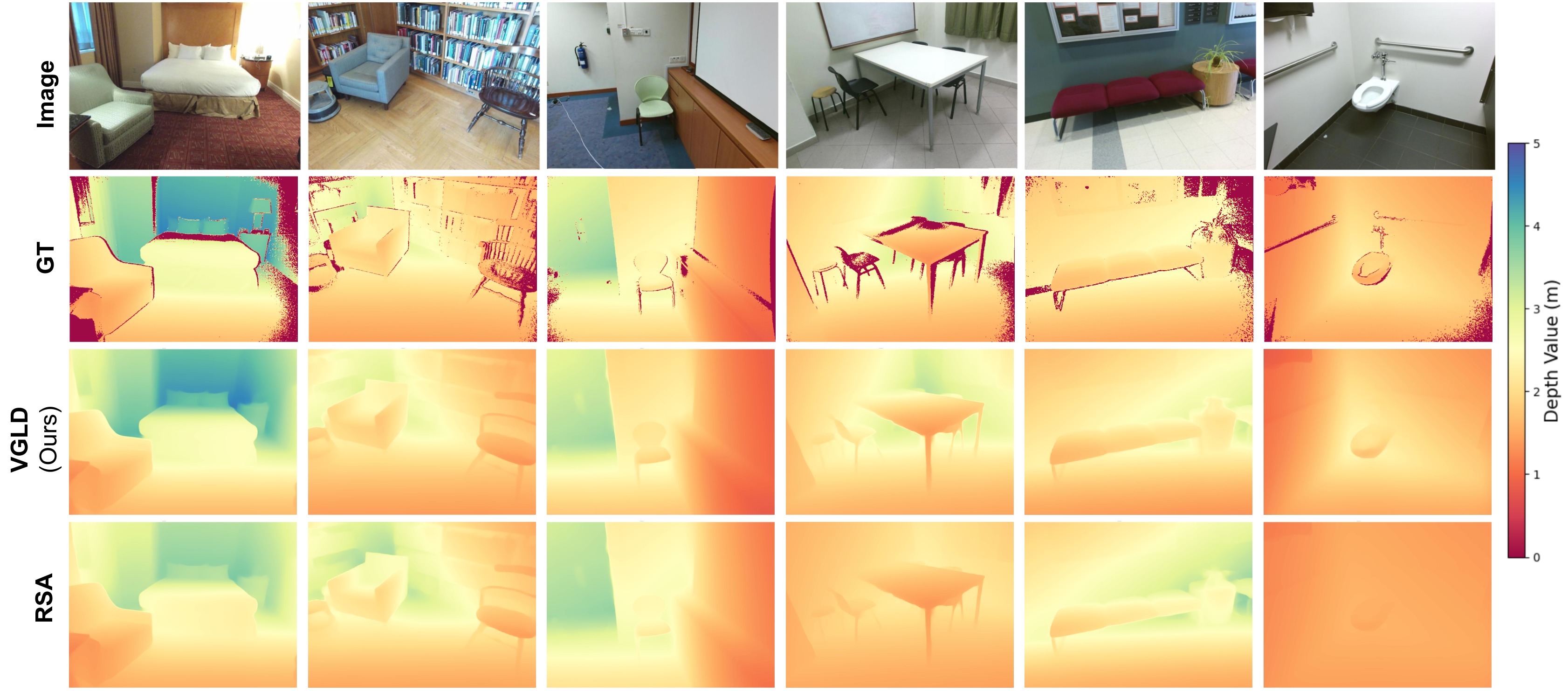}
\caption{Zero-shot generalization on the SUN-RGBD dataset(Indoor). The models are evaluated without any fine-tuning. Benefiting from robust scale prediction, our VGLD method produces depth maps that are significantly closer to the ground truth compared to RSA. }
\label{fig_5}
\end{figure*}

\begin{figure*}[t]
\centering
\includegraphics[width=1\textwidth]{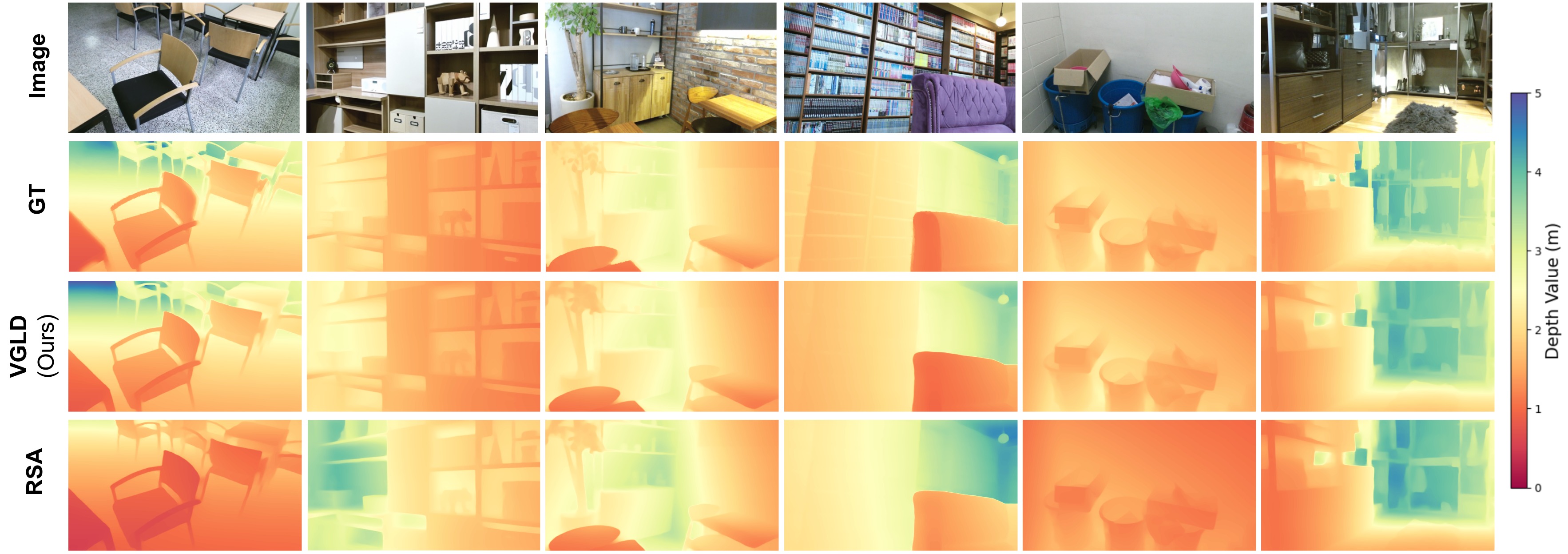}
\caption{Zero-shot generalization on the DIML Indoor dataset(Indoor). The models are evaluated without any fine-tuning. Benefiting from robust scale prediction, our VGLD method produces depth maps that are significantly closer to the ground truth compared to RSA. }
\label{fig_6}
\end{figure*}

\begin{figure*}[t]
\centering
\includegraphics[width=1\textwidth]{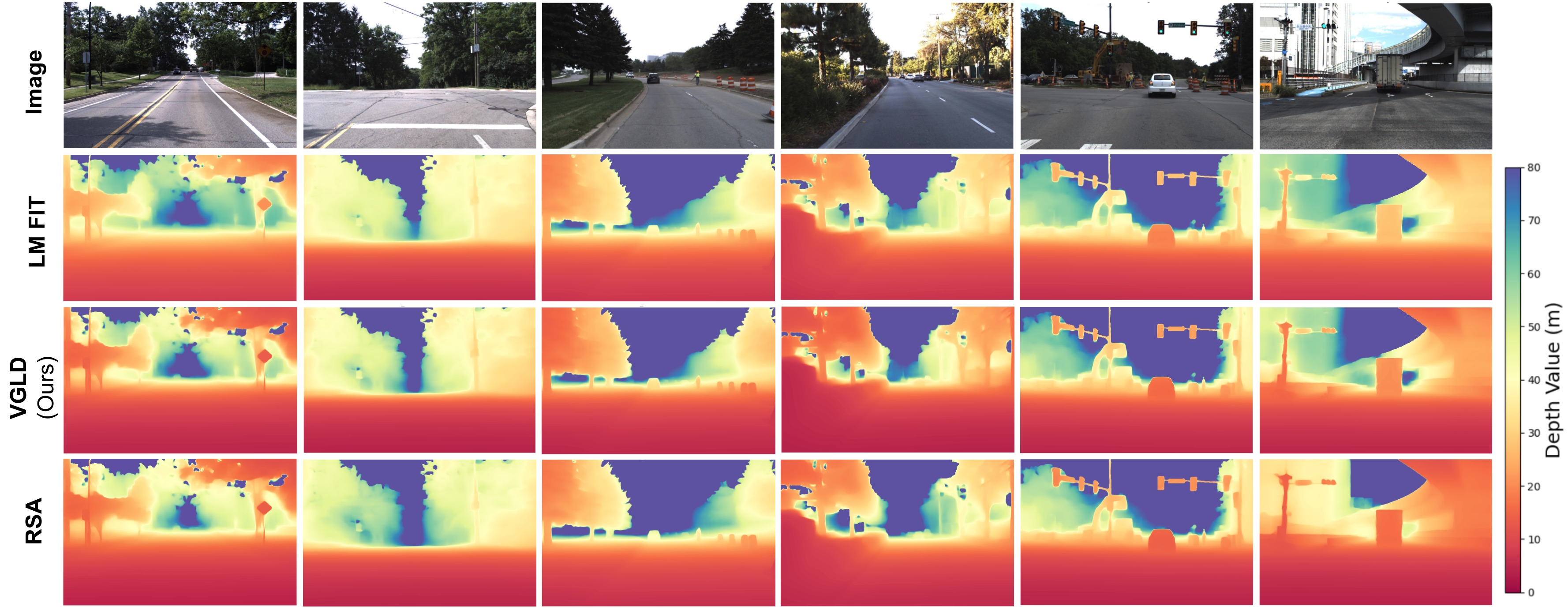}
\caption{Zero-shot generalization on the DDAD dataset(Outdoor). The models are evaluated without any fine-tuning. Benefiting from robust scale prediction, our VGLD method produces depth maps that are significantly closer to the ground truth compared to RSA. 
Note that due to the sparse ground truth depth maps in the DDAD dataset, the visualization quality is poor. 
Therefore, LM Fit is used as a substitute for the ground truth depth map in the visualizations.}
\label{fig_7}
\end{figure*}

Benefiting from the smaller domain gap of language descriptions across diverse scenes\cite{zeng2024wordepth, zeng2024rsa} and the ability of corresponding images to accurately indicate domain context, 
we conduct a zero-shot transfer experiment to demonstrate the generalization capability of VGLD.
We evaluate the models on the SUN-RGBD\cite{song2015sun} , DIML Indoor\cite{cho2021diml}, and DDAD\cite{guizilini20203d} datasets without any fine-tuning. 
As shown in Figure \ref{fig_5}, Figure \ref{fig_6}, Figure \ref{fig_7} (qualitative results) and Table \ref{tab:3}, Table \ref{tab:4}, Table \ref{tab:10} (quantitative results), VGLD consistently outperforms baseline methods and produces results that closely match those fitted by the LM method. This demonstrates that, under visual guidance, VGLD maintains stable scalars estimation and exhibits enhanced generalization capabilities. 
Note that all zero-shot experiments are conducted using the VGLD-NK-TCI model built upon the DAV1-vits RDE backbone.

\subsection{Effect of the initial seeds}
To ensure the robustness of our training and verify that the results are not due to random initialization, we trained the model using three different random seeds. 
As illustrated in Figure \ref{fig:diff_seed}, the resulting error bars indicate that variations due to different seeds are minimal, with nearly zero deviation.

\subsection{Prompts for Natural Text Generation}
To generate natural and semantically rich image descriptions—rather than relying on fixed prompt templates—we employ two vision-language models: LLaVA-v1.6-Vicuna-7B and LLaVA-v1.6-Mistral-7B\cite{jia2022visual}. 
To ensure diversity in the generated captions, each model is prompted using six distinct instruction templates. These prompt templates are listed in Table \ref{tab:prompt text}.

\begin{table*}[h]
\centering
\setlength{\tabcolsep}{1mm} 
{\fontsize{9}{11}\selectfont
\begin{tabular}{l|l|ccccc|ccc}
\hline
Models & Methods & Abs Rel $\downarrow$ & sq\_rel $\downarrow$& RMSE $\downarrow$ & $\text{RMSE}_\text{log}$ $\downarrow$ & log$_{10}$ $\downarrow$ & D1 $\uparrow$ & D2 $\uparrow$ & D3 $\uparrow$ \\
\hline
ZoeDept\cite{bhat2023zoedepth}h & \multirow{3}{*}{\centering robust depth estimation } & 0.077 & -- & 0.277 & -- & 0.033 & 0.953 & 0.995 & 0.999 \\
ZeroDepth\cite{guizilini2023towards} & & 0.074 & -- & 0.269 & -- & 0.103 & 0.954 & 0.995 & 1.000 \\ 
Metric3Dv2\cite{hu2024metric3d} & & 0.047 & -- & 0.183 & -- & 0.020 & 0.989 & 0.998 & 1.000 \\
\hline
\multirow{10}{*}{MiDas-1\cite{birkl2023midas}} 

& Least Squares & 0.121 & 0.073 & 0.388 & 0.338 & 0.068 & 0.866 & 0.959 & 0.978 \\  
& Levenberg Marquardt & 0.056 & 0.021 & 0.218 & 0.080 & 0.024 & 0.969 & 0.995 & 0.998 \\
\cline{2-10}
& RSA-N\cite{zeng2024rsa} & 0.171 & -- & 0.569 & -- & 0.072 & 0.731 & 0.955 & 0.993 \\
& RSA-NK\cite{zeng2024rsa} & 0.168 & -- & 0.561 & -- & 0.071 & 0.737 & 0.959 & 0.993 \\
& VGLD-N-T (Ours) & 0.158 & 0.113 & 0.529 & 0.181 & 0.068 & 0.758 & 0.965 & 0.994 \\
& VGLD-N-I (Ours) & 0.121 & \underbar{0.068} & \underbar{0.423} & 0.146 & 0.053 & 0.860 & \textbf{0.985} & \textbf{0.998} \\
& VGLD-N-TCI (Ours) & \textbf{0.119} & \textbf{0.067} & \textbf{0.414} & \textbf{0.142} & \textbf{0.051} & \textbf{0.867} & \underbar{0.984} & \textbf{0.998} \\
& VGLD-NK-T (Ours) & 0.159 & 0.113 & 0.526 & 0.178 & 0.067 & 0.751 & 0.971 & \underbar{0.995} \\
& VGLD-NK-I (Ours) & 0.123 & 0.070 & 0.426 & 0.147 & 0.053 & 0.855 & 0.982 & \textbf{0.998} \\
& VGLD-NK-TCI (Ours) & \underbar{0.120} & \underbar{0.068} & \textbf{0.414} & \underbar{0.143} & \underbar{0.052} & \underbar{0.863} & \underbar{0.984} & \textbf{0.998} \\
\hline
\multirow{8}{*}{MiDas-2\cite{ranftl2020towards}}
& Least Squares & 0.130 & 0.085 & 0.421 & 0.286 & 0.066 & 0.845 & 0.956 & 0.980 \\

& Levenberg Marquardt & 0.094 & 0.049 & 0.330 & 0.122 & 0.039 & 0.916 & 0.985 & 0.997 \\
\cline{2-10} 
& VGLD-N-T (Ours) & 0.180 & 0.140 & 0.596 & 0.212 & 0.078 & 0.688 & 0.946 & 0.990 \\
& VGLD-N-I (Ours) & \underbar{0.154} & \underbar{0.106} & 0.524 & 0.186 & 0.067 & 0.775 & 0.960 & \textbf{0.993} \\
& VGLD-N-TCI (Ours) & \textbf{0.151} & \textbf{0.104} & \textbf{0.507} & \textbf{0.181} & \textbf{0.064} & \textbf{0.789} & \textbf{0.964} & \textbf{0.993} \\
& VGLD-NK-T (Ours) & 0.182 & 0.147 & 0.615 & 0.217 & 0.080 & 0.682 & 0.939 & 0.989 \\
& VGLD-NK-I (Ours) & 0.155 & 0.108 & 0.520 & 0.185 & 0.066 & 0.776 & \underbar{0.961} & \underbar{0.992} \\
& VGLD-NK-TCI (Ours) & \textbf{0.151} & \textbf{0.104} & \underbar{0.513} & \underbar{0.183} & \underbar{0.065} & \underbar{0.780} & \textbf{0.964} & \textbf{0.993} \\
\hline

\multirow{8}{*}{DAV2-vits\cite{yang2024depthv2}}
& Least Squares & 0.122 & 0.074 & 0.392 & 0.362 & 0.070 & 0.866 & 0.959 & 0.977 \\

& Levenberg Marquardt & 0.052 & 0.021 & 0.209 & 0.077 & 0.022 & 0.969 & 0.992 & 0.998 \\
\cline{2-10} 
& VGLD-N-T (Ours) & 0.163 & 0.119 & 0.546 & 0.191 & 0.073 & 0.713 & 0.964 & \underbar{0.994} \\
& VGLD-N-I (Ours) & 0.128 & \underbar{0.074} & \underbar{0.433} & 0.154 & \underbar{0.057} & \underbar{0.830} & \underbar{0.983} & \textbf{0.995} \\
& VGLD-N-TCI (Ours) & \textbf{0.125} & \textbf{0.073} & \textbf{0.423} & \textbf{0.152} & \textbf{0.055} & \textbf{0.842} & \textbf{0.984} & \textbf{0.995} \\
& VGLD-NK-T (Ours) & 0.161 & 0.115 & 0.539 & 0.189 & 0.073 & 0.714 & 0.967 & \underbar{0.994} \\
& VGLD-NK-I (Ours) & \underbar{0.127} & \underbar{0.074} & 0.436 & \underbar{0.155} & \underbar{0.057} & 0.835 & 0.982 & \textbf{0.995} \\
& VGLD-NK-TCI (Ours) & \underbar{0.127} & \underbar{0.074} & 0.434 & \underbar{0.155} & \underbar{0.057} & 0.835 & 0.981 & \textbf{0.995} \\
\hline

\multirow{10}{*}{DAV1-vits\cite{yang2024depth}}
& Least Squares & 0.121 & 0.075 & 0.397 & 0.327 & 0.067 & 0.863 & 0.959 & 0.979 \\

& Levenberg Marquardt & 0.057 & 0.022 & 0.230 & 0.081 & 0.024 & 0.967 & 0.995 & 0.999 \\
\cline{2-10} 
& RSA-N\cite{zeng2024rsa} & 0.147 & -- & 0.484 & -- & 0.065 & 0.775 & 0.975 & 0.997 \\
& RSA-NK\cite{zeng2024rsa} & 0.148 & -- & 0.498 & -- & 0.065 & 0.776 & 0.974 & 0.996 \\
& VGLD-N-T (Ours) & 0.145 & 0.094 & 0.496 & 0.170 & 0.064 & 0.792 & 0.974 & 0.997 \\
& VGLD-N-I (Ours) & 0.115 & 0.061 & 0.405 & 0.141 & 0.051 & 0.872 & \underbar{0.987} & \underbar{0.998} \\
& VGLD-N-TCI (Ours) & \textbf{0.112} & \textbf{0.058} & \textbf{0.390} & \textbf{0.135} & \underbar{0.049} & \textbf{0.887} & \textbf{0.988} & \underbar{0.998} \\
& VGLD-NK-T (Ours) & 0.142 & 0.089 & 0.483 & 0.168 & 0.063 & 0.787 & 0.979 & 0.997 \\
& VGLD-NK-I (Ours) & \underbar{0.114} & 0.061 & 0.404 & \underbar{0.138} & 0.050 & 0.880 & \underbar{0.987} & \textbf{0.999} \\
& VGLD-NK-TCI (Ours) & \textbf{0.112} & \underbar{0.059} & \underbar{0.392} & \textbf{0.135} & \textbf{0.048} & \underbar{0.883} & \textbf{0.988} & \textbf{0.999} \\
\bottomrule
\end{tabular}}
\caption{More detailed quantitative depth comparison on the NYUv2 dataset. Best results are in \textbf{bold}, second best are \underline{underlined}.
}
\label{tab:Appendix_nyu}
\end{table*}

\begin{table*}[h]
\centering
\setlength{\tabcolsep}{1mm} 
{\fontsize{9}{11}\selectfont
\begin{tabular}{l|l|ccccc|ccc}
\hline
Models & Methods & Abs Rel $\downarrow$ & sq\_rel $\downarrow$& RMSE $\downarrow$ & $\text{RMSE}_\text{log}$ $\downarrow$ & log$_{10}$ $\downarrow$ & D1 $\uparrow$ & D2 $\uparrow$ & D3 $\uparrow$ \\
\hline
ZoeDepth\cite{bhat2023zoedepth} & \multirow{3}{*}{\centering robust depth estimation } & 0.054 & -- & 2.281 & 0.082 & -- & 0.971 & 0.996 & 0.999 \\
ZeroDepth\cite{guizilini2023towards} & & 0.053 & -- & 2.087 & 0.083 & -- & 0.968 & 0.995 & 0.999 \\
Metric3Dv2\cite{hu2024metric3d} & & 0.044 & -- & 1.985 & 0.064 & -- & 0.985 & 0.998 & 0.999 \\
\hline

\multirow{10}{*}{MiDas-1\cite{birkl2023midas}} 
& Least Squares & 0.333 & 2.094 & 6.901 & 1.731 & 0.293 & 0.408 & 0.790 & 0.879 \\

& Levenberg Marquardt & 0.091 & 0.425 & 3.373 & 0.127 & 0.038 & 0.925 & 0.987 & 0.996 \\
\cline{2-10} 
& RSA-K\cite{zeng2024rsa} & 0.163 & -- & 4.082 & 0.185 & -- & 0.798 & 0.948 & 0.981 \\
& RSA-NK\cite{zeng2024rsa} & 0.160 & -- & 4.232 & 0.194 & -- & 0.782 & 0.946 & 0.980 \\
& VGLD-K-T(Ours) & 0.133 & 0.608 & 3.755 & 0.162 & \underbar{0.056} & 0.854 & 0.975 & 0.993 \\
& VGLD-K-I(Ours) & \textbf{0.120} & 0.526 & 3.668 & 0.152 & \textbf{0.051} & 0.868 & \underbar{0.979} & \underbar{0.995} \\
& VGLD-K-TCI(Ours) & \textbf{0.120} & \textbf{0.523} & 3.598 & \underbar{0.151} & \textbf{0.051} & \underbar{0.871} & \textbf{0.980} & \textbf{0.996} \\
& VGLD-NK-T(Ours) & 0.130 & 0.568 & 3.744 & 0.161 & \underbar{0.056} & 0.844 & 0.975 & \underbar{0.995} \\
& VGLD-NK-I(Ours) & \underbar{0.122} & 0.543 & \underbar{3.574} & \underbar{0.151} & \textbf{0.051} & 0.868 & \underbar{0.979} & \underbar{0.995} \\
& VGLD-NK-TCI(Ours) & \textbf{0.120} & \underbar{0.528} & \textbf{3.559} & \textbf{0.150} & \textbf{0.051} & \textbf{0.874} & \textbf{0.980} & \textbf{0.996} \\
\hline

\multirow{8}{*}{MiDas-2\cite{ranftl2020towards}}
& Least Squares & 0.336 & 2.172 & 6.925 & 1.658 & 0.283 & 0.421 & 0.778 & 0.876 \\

& Levenberg Marquardt & 0.155 & 0.770 & 4.190 & 0.185 & 0.062 & 0.809 & 0.966 & 0.990 \\
\cline{2-10} 
& VGLD-K-T(Ours) & 0.194 & 1.290 & 5.030 & 0.225 & 0.079 & 0.709 & 0.930 & 0.981 \\
& VGLD-K-I(Ours) & 0.183 & 1.154 & 4.842 & 0.215 & 0.075 & 0.733 & \underbar{0.942} & \underbar{0.983} \\
& VGLD-K-TCI(Ours) & \textbf{0.178} & \textbf{1.146} & \underbar{4.806} & \textbf{0.210} & \textbf{0.073} & \textbf{0.748} & \underbar{0.942} & \textbf{0.984} \\
& VGLD-NK-T(Ours) & 0.191 & 1.260 & 4.994 & 0.221 & 0.078 & 0.723 & 0.932 & 0.981 \\
& VGLD-NK-I(Ours) & 0.184 & 1.179 & 4.808 & 0.213 & \underbar{0.074} & \underbar{0.740} & 0.938 & \textbf{0.984} \\
& VGLD-NK-TCI(Ours) & \underbar{0.180} & \underbar{1.158} & \textbf{4.804} & \underbar{0.212} & \underbar{0.074} & 0.737 & \textbf{0.943} & \textbf{0.984} \\
\hline

\multirow{8}{*}{DAV2-vits\cite{yang2024depthv2}}
& Least Squares & 0.330 & 2.053 & 6.737 & 1.729 & 0.292 & 0.423 & 0.790 & 0.877 \\

& Levenberg Marquardt & 0.103 & 0.454 & 3.277 & 0.135 & 0.042 & 0.919 & 0.987 & 0.997 \\
\cline{2-10} 
& VGLD-K-T(Ours)    & {0.166} & {0.822} & {4.189} & {0.190} & {0.070} & {0.756} & {0.953} & {0.992} \\
& VGLD-K-I(Ours)    & {0.154} & {0.698} & {4.219} & {0.187} & {0.067} & {0.756} & {0.966} & \underbar{0.995} \\
& VGLD-K-TCI(Ours)  & \textbf{0.152} & \textbf{0.657} & \textbf{3.872} & \textbf{0.179} & \textbf{0.065} & \textbf{0.779} & \underbar{0.972} & \textbf{0.996} \\
& VGLD-NK-T(Ours)   & {0.164} & {0.786} & {4.287} & {0.193} & {0.070} & {0.752} & {0.955} & {0.993} \\
& VGLD-NK-I(Ours)   & {0.160} & {0.748} & {4.031} & {0.187} & {0.069} & {0.761} & {0.965} & \underbar{0.995} \\
& VGLD-NK-TCI(Ours) & \underbar{0.153} & \underbar{0.695} & \underbar{3.980} & \underbar{0.180} & \underbar{0.066} & \underbar{0.772} & \textbf{0.973} & \textbf{0.996} \\
\hline

\multirow{10}{*}{DAV1-vits\cite{yang2024depth}}
& Least Squares & 0.331 & 2.078 & 6.772 & 1.714 & 0.291 & 0.423 & 0.786 & 0.875 \\

& Levenberg Marquardt & 0.112 & 0.495 & 3.375 & 0.142 & 0.045 & 0.897 & 0.986 & 0.997 \\
\cline{2-10} 
& RSA-K\cite{zeng2024rsa} & {0.160}   & -- & {4.437} & {0.189} & -- & {0.780} & {0.958} & {0.988} \\
& RSA-NK\cite{zeng2024rsa} & {0.158}  & -- & {4.457} & {0.179} & -- & {0.786} & {0.967} & {0.987} \\
& VGLD-K-T(Ours)    & {0.151} & {0.747} & {4.354} & {0.186} & {0.066} & {0.773} & {0.963} & {0.994} \\
& VGLD-K-I(Ours)    & {0.144} & \underbar{0.646} & \underbar{4.074} & {0.178} & {0.063} & {0.790} & \underbar{0.975} & \underbar{0.996} \\
& VGLD-K-TCI(Ours)  & \underbar{0.140} & {0.686} & {4.081} & \underbar{0.172} & \underbar{0.061} & {0.807} & \underbar{0.975} & \underbar{0.996} \\
& VGLD-NK-T(Ours)   & {0.148} & {0.728} & {4.293} & {0.183} & {0.065} & {0.781} & {0.966} & {0.995} \\
& VGLD-NK-I(Ours)   & {0.142} & {0.759} & {4.151} & \underbar{0.172} & \underbar{0.061} & \underbar{0.814} & \underbar{0.975} & \underbar{0.996} \\
& VGLD-NK-TCI(Ours) & \textbf{0.136} & \textbf{0.632} & \textbf{4.008} & \textbf{0.169} & \textbf{0.059} & \textbf{0.816} & \textbf{0.977} & \textbf{0.997} \\
\bottomrule
\end{tabular}
}
\caption{More detailed quantitative depth comparison on the KITTI dataset. Best results are in \textbf{bold}, second best are \underline{underlined}.}\label{tab:Appendix_kitti}
\end{table*}


\begin{table*}[h]
\centering
\setlength{\tabcolsep}{2mm} 
{\fontsize{9}{11}\selectfont
\begin{tabular}{|c|c|c|c|c|c|c|c|c|c|}
\hline
\textbf{Idx} & \textbf{Text-idx} & \textbf{Method} & \textbf{Abs Rel $\downarrow$} & \textbf{RMSE $\downarrow$} & \textbf{D1 $\uparrow$} & \textbf{pred\_shift} & \textbf{LM\_shift} & \textbf{pred\_scale} & \textbf{LM\_scale} \\ 
\hline
\multirow{6}{*}{1} & \multirow{2}{*}{Text-1} & RSA\cite{zeng2024rsa}  & {0.210} & {0.689} & {0.263} & {1.255} &                              & {1.032} &       \\
                    &                         & VGLD & \textbf{0.080} & \textbf{0.240} & \textbf{0.987} & {1.207} &                              & {1.020} &       \\ \cline{2-7}  \cline{9-9} 
                    & \multirow{2}{*}{Text-2} & RSA\cite{zeng2024rsa}  & {0.110} & {0.367} & {0.995} & {1.220} &                              & {1.028} &       \\
                    &                         & VGLD & \textbf{0.065} & \textbf{0.271} & \textbf{0.999} & {1.220} &                              & {1.284} &       \\ \cline{2-7}  \cline{9-9} 
                    & \multirow{2}{*}{Text-3} & RSA\cite{zeng2024rsa}  & {0.054} & \textbf{0.216} & \textbf{0.997} & {1.210} &                              & {1.026} &       \\ 
                    &                         & VGLD & \textbf{0.052} & {0.232} & \textbf{0.997} & {1.216} & \multirow{-6}{*}{1.193}      & {1.025} &  \multirow{-6}{*}{1.026}     \\
\hline
\multirow{6}{*}{2} & \multirow{2}{*}{Text-1} & RSA\cite{zeng2024rsa}  & {0.089} & {0.344} & {0.961} & {1.202} &                         & {1.032} &       \\
                    &                         & VGLD & \textbf{0.073} & \textbf{0.271} & \textbf{0.962} & {1.185} &                         & {1.034} &       \\ \cline{2-7}  \cline{9-9} 
                    & \multirow{2}{*}{Text-2} & RSA\cite{zeng2024rsa}  & {0.069} & \textbf{0.256} & {0.956} & {1.214} &                         & {1.030} &       \\
                    &                         & VGLD & \textbf{0.063} & {0.272} & \textbf{0.962} & {1.213} &                         & {1.037} &       \\ \cline{2-7}  \cline{9-9} 
                    & \multirow{2}{*}{Text-3} & RSA\cite{zeng2024rsa}  & {0.064} & {0.296} & {0.947} & {1.218} &                         & {1.036} &       \\
                    &                         & VGLD & \textbf{0.062} & \textbf{0.280} & \textbf{0.954} & {1.228} & \multirow{-6}{*}{1.210} & {1.034} &  \multirow{-6}{*}{1.033} \\
\hline
\multirow{6}{*}{3} & \multirow{2}{*}{Text-1} & RSA\cite{zeng2024rsa}  & {0.251} & {0.868} & {0.138} & {1.331} &                         & {1.042} &       \\
                    &                         & VGLD & \textbf{0.147} & \textbf{0.433} & \textbf{0.920} & {1.254} &                         & {1.041} &       \\ \cline{2-7}  \cline{9-9} 
                    & \multirow{2}{*}{Text-2} & RSA\cite{zeng2024rsa}  & {0.055} & {0.240} & {0.993} & {1.199} &                         & {1.035} &       \\
                    &                         & VGLD & \textbf{0.054} & \textbf{0.170} & \textbf{0.994} & {1.205} &                         & {1.038} &       \\ \cline{2-7}  \cline{9-9} 
                    & \multirow{2}{*}{Text-3} & RSA\cite{zeng2024rsa}  & {0.058} & {0.154} & \textbf{0.994} & {1.212} &                         & {1.039} &       \\
                    &                         & VGLD & \textbf{0.055} & \textbf{0.138} & \textbf{0.994} & {1.199} & \multirow{-6}{*}{1.218} & {1.035} & \multirow{-6}{*}{1.034}     \\
\hline
\end{tabular}
}
\caption{Quantitative results on the NYUv2 dataset comparing VGLD and RSA in response to different textual descriptions. The LM\_shift and LM\_scale represent scalars values fitted using the Levenberg-Marquardt method.
Best results are in \textbf{bold}.}
\label{tab:diff nyu}
\end{table*}

\begin{table*}[h]
\centering
\setlength{\tabcolsep}{3mm} 
{\fontsize{10}{12}\selectfont
\begin{tabular}{|c|c|p{11cm}|}
\hline
\textbf{Idx} & \textbf{Texts-idx} & \textbf{Text Description} \\ 
\hline
\multirow{3}{*}{1}  &  Text-1 & A man is standing in a doorway, looking at a bed with a striped comforter. \\ \cline{2-3}
                    &  Text-2 & The bed is positioned in the corner of the room, with a man standing in the doorway, and a fish tank nearby. \\ \cline{2-3}
                    &  Text-3 & A man stands in a doorway, looking into a bedroom with a large bed, a wooden dresser, and a fish tank.  \\
\hline
\multirow{3}{*}{2}  &  Text-1 & The image shows a classroom with a play area, a table with chairs, and a sink. \\ \cline{2-3}
                    &  Text-2 & The image shows a classroom with a table, chairs, and a sink, all situated near a wall with bulletin boards and a window.                    \\ \cline{2-3}
                    &  Text-3 & The image shows a classroom with a table, chairs, a sink, a bulletin board, a bookshelf, a window, and a rug.                    \\
\hline
\multirow{3}{*}{3}  &  Text-1 & The image shows a red couch with towels hanging over the back, a flat screen television, and a framed jersey on the wall.\\ \cline{2-3}
                    &  Text-2 & The image shows a red couch with a pink towel and a blue towel on it, positioned in front of a television with a framed jersey on the wall behind it. \\ \cline{2-3}
                    &  Text-3 & The image shows a living room with a red couch, a flat screen TV, a framed jersey, and a guitar. \\
\hline
\end{tabular}
}
\caption{The table shows three distinct textual descriptions provided for each image in the NYUv2 dataset, used as linguistic inputs for evaluating model sensitivity.}
\label{tab:nyu text}
\end{table*}
  
\begin{table*}[h]
\centering
\setlength{\tabcolsep}{2mm} 
{\fontsize{9}{11}\selectfont
\begin{tabular}{|c|c|c|c|c|c|c|c|c|c|}
\hline
\textbf{Idx} & \textbf{Text-idx} & \textbf{Method} & \textbf{Abs Rel $\downarrow$} & \textbf{RMSE $\downarrow$} & \textbf{D1 $\uparrow$} & \textbf{pred\_shift} & \textbf{LM\_shift} & \textbf{pred\_scale} & \textbf{LM\_scale} \\ 
\hline
\multirow{6}{*}{1} & \multirow{2}{*}{Text-1}  & RSA\cite{zeng2024rsa}  & {0.097} & {4.060} & {0.926} & {1.005} & & {1.011} &  \\ 
                    &                         & VGLD & \textbf{0.075} & \textbf{3.562} & \textbf{0.949} & {1.004} &       & {1.001} &       \\ \cline{2-7}  \cline{9-9} 
                    & \multirow{2}{*}{Text-2} & RSA\cite{zeng2024rsa}  & {0.084} & {4.251} & {0.923} & {1.007} &       & {1.010} &       \\
                    &                         & VGLD & \textbf{0.077} & \textbf{3.067} & \textbf{0.949} & {1.004} &       & {1.010} &       \\ \cline{2-7}  \cline{9-9}  
                    & \multirow{2}{*}{Text-3} & RSA\cite{zeng2024rsa}  & {0.072} & {3.140} & \textbf{0.952} & {1.004} &       & {1.010} &       \\
                    &                         & VGLD & \textbf{0.068} & \textbf{3.088} & {0.951} & {1.004} & \multirow{-6}{*}{1.003} & {1.010} & \multirow{-6}{*}{1.010} \\
\hline
\multirow{6}{*}{2} & \multirow{2}{*}{Text-1}  & RSA\cite{zeng2024rsa}  & {0.108} & {2.327} & {0.905} & {1.009} &       & {1.014} & \\
                    &                         & VGLD & \textbf{0.063} & \textbf{1.887} & \textbf{0.984} & {1.135} &       & {1.015} &       \\ \cline{2-7}  \cline{9-9} 
                    & \multirow{2}{*}{Text-2} & RSA\cite{zeng2024rsa}  & {0.281} & {5.341} & {0.537} & {1.006} &       & {1.013} &       \\
                    &                         & VGLD & \textbf{0.099} & \textbf{2.144} & \textbf{0.915} & {1.010} &       & {1.014} &       \\ \cline{2-7}  \cline{9-9} 
                    & \multirow{2}{*}{Text-3} & RSA\cite{zeng2024rsa}  & {0.109} & {2.157} & {0.906} & {1.011} &       & {1.014} &       \\
                    &                         & VGLD & \textbf{0.073} & \textbf{1.861} & \textbf{0.952} & {1.011} & \multirow{-6}{*}{1.009} & {1.015} & \multirow{-6}{*}{1.017} \\
\hline
\multirow{6}{*}{3} & \multirow{2}{*}{Text-1} & RSA\cite{zeng2024rsa}  & {0.268} & {6.526} & {0.740} & {1.004}  &       & {1.009} &  \\
                    &                         & VGLD & \textbf{0.119} & \textbf{2.516} & \textbf{0.919} & {1.009} &       & {1.010} &       \\ \cline{2-7}  \cline{9-9} 
                    & \multirow{2}{*}{Text-2} & RSA\cite{zeng2024rsa}  & {0.171} & {4.479} & {0.854} & {1.005} &       & {1.010} &       \\
                    &                         & VGLD & \textbf{0.077} & \textbf{2.287} & \textbf{0.942} & {1.008} &       & {1.010} &       \\ \cline{2-7}  \cline{9-9} 
                    & \multirow{2}{*}{Text-3} & RSA\cite{zeng2024rsa}  & {0.081} & {2.421} & {0.938} & {1.007} &       & {1.010} &       \\
                    &                         & VGLD & \textbf{0.062} & \textbf{2.236} & \textbf{0.953} & {1.009} & \multirow{-6}{*}{1.008} & {1.011} &  \multirow{-6}{*}{1.011}     \\
\hline
\end{tabular}
}
\caption{Quantitative results on the KITTI dataset comparing VGLD and RSA in response to different textual descriptions. The LM\_shift and LM\_scale represent scalars values fitted using the Levenberg-Marquardt method.
Best results are in \textbf{bold}.}
\label{tab:diff kitti}
\end{table*}

\begin{table*}[h]
\centering
\setlength{\tabcolsep}{3mm} 
{\fontsize{10}{12}\selectfont
\begin{tabular}{|c|c|p{11cm}|}
\hline
\textbf{Idx} & \textbf{Text-idx} & \textbf{Text Description} \\ 
\hline
\multirow{3}{*}{1}  &  Text-1 & The image shows a narrow city street lined with parked cars and buildings on both sides. \\ \cline{2-3}
                    &  Text-2 & The image shows a narrow street lined with parked cars and buildings, with a clear sky overhead.      \\ \cline{2-3}
                    &  Text-3 & The image shows a narrow street with parked cars on both sides, leading towards a building with a red awning. \\
\hline
\multirow{3}{*}{2}  &  Text-1 & The image shows a narrow alleyway with a white gate at the end, a bridge overhead, and a hillside on one side. \\ \cline{2-3}
                    &  Text-2 & The image shows a narrow alleyway with a white gate, a fence, a building, a bridge, and a sign, all situated in close proximity to each other. \\ \cline{2-3}
                    &  Text-3 & A narrow alleyway with a white gate, a fence, a building, a bridge, a tree, a sign, and a hill. \\
\hline
\multirow{3}{*}{3}  &  Text-1 & The image captures a lively street scene with people walking and riding bicycles, shops and buildings lining the street, and a clear blue sky overhead. \\ \cline{2-3}
                    &  Text-2 & The image shows a narrow street in a European city, with buildings on both sides, a pedestrian walkway in the middle, and people walking and biking on the street. \\ \cline{2-3}
                    &  Text-3 & The image shows a bustling city street with people walking and riding bicycles, shops and buildings lining the street, and a clear blue sky overhead. \\
\hline
\end{tabular}
}
\caption{The table shows three distinct textual descriptions provided for each image in the KITTI dataset, used as linguistic inputs for evaluating model sensitivity.}
\label{tab:kitti text}
\end{table*}

\begin{table*}[h]
\centering
\setlength{\tabcolsep}{1mm} 
{\fontsize{9}{11}\selectfont
\begin{tabular}{l|l|ccccc|ccc}
\hline
\multirow{2}{*}{\textbf{RDE Model}} & \multirow{2}{*}{\textbf{Method}} & \multicolumn{5}{c|}{\textbf{Lower is better}} & \multicolumn{3}{c}{\textbf{Higher is better}} \\  
& & \textbf{Abs Rel $\downarrow$} & \textbf{sq\_rel $\downarrow$} & \textbf{RMSE $\downarrow$} & \textbf{$\text{RMSE}_\text{log}$ $\downarrow$} & \textbf{log$_{10}$ $\downarrow$} & \textbf{D1 $\uparrow$} & \textbf{D2 $\uparrow$} & \textbf{D3 $\uparrow$} \\
\hline
ZoeDepth\cite{bhat2023zoedepth} & \multirow{2}{*}{\centering robust depth estimation$\dagger$} & 0.123 & -- & 0.356 & -- & 0.053 & 0.856 & 0.979 & 0.995 \\
ScaleDepth\cite{zhu2024scaledepth} & & 0.129 & -- & 0.359 & -- & -- & 0.866 & -- & -- \\
\hline

\multirow{6}{*}{MiDas-1\cite{birkl2023midas}} 
& Least Squares & 0.197 & 0.418 & 0.346 & 0.278 & 0.061 & 0.873 & 0.964 & 0.981 \\

& Levenberg Marquardt & 0.158 & 0.440 & 0.252 & 0.116 & 0.032 & 0.950 & 0.988 & 0.995 \\
\cline{2-10} 
& RSA-NK\cite{zeng2024rsa}            & {0.299} & \textbf{0.589} & {0.575} & {0.251} & {0.094} & {0.615} & {0.900} & {0.977} \\
& VGLD-NK-T(Ours)   & {0.318} & {0.647} & {0.566} & \underbar{0.242} & \underbar{0.089} & \underbar{0.643} & {0.914} & \underbar{0.980} \\
& VGLD-NK-I(Ours)   & \textbf{0.259} & \underbar{0.595} & \underbar{0.468} & \textbf{0.202} & \textbf{0.071} & \textbf{0.751} & \underbar{0.957} & \textbf{0.991} \\
& VGLD-NK-TCI(Ours) & \underbar{0.262} & {0.628} & \textbf{0.467} & \textbf{0.202} & \textbf{0.071} & \textbf{0.751} & \textbf{0.959} & \textbf{0.991} \\
\hline

\multirow{5}{*}{MiDas-2\cite{ranftl2020towards}}
& Least Squares & 0.203 & 0.419 & 0.365 & 0.272 & 0.062 & 0.860 & 0.961 & 0.981 \\

& Levenberg Marquardt & 0.173 & 0.438 & 0.291 & 0.132 & 0.039 & 0.926 & 0.984 & 0.994 \\
\cline{2-10} 
& VGLD-NK-T(Ours)   & {0.316} & {0.795} & {0.597} & {0.249} & \underbar{0.090} & \underbar{0.639} & {0.908} & {0.978} \\
& VGLD-NK-I(Ours)   & \underbar{0.288} & \underbar{0.688} & \underbar{0.552} & \underbar{0.246} & \underbar{0.090} & {0.627} & \underbar{0.922} & \underbar{0.984} \\
& VGLD-NK-TCI(Ours) & \textbf{0.275} & \textbf{0.670} & \textbf{0.513} & \textbf{0.225} & \textbf{0.080} & \textbf{0.694} & \textbf{0.941} & \textbf{0.987} \\
\hline

\multirow{5}{*}{DAV2-vits\cite{yang2024depthv2}}
& Least Squares & 0.194 & 0.418 & 0.337 & 0.305 & 0.062 & 0.880 & 0.963 & 0.980 \\

& Levenberg Marquardt & 0.146 & 0.439 & 0.224 & 0.103 & 0.027 & 0.961 & 0.989 & 0.995 \\
\cline{2-10} 
& VGLD-NK-T(Ours)   & {0.304} & {0.742} & {0.564} & \underbar{0.236} & \underbar{0.089} & \underbar{0.644} & {0.920} & {0.983} \\
& VGLD-NK-I(Ours)   & \underbar{0.273} & \underbar{0.564} & \underbar{0.535} & \underbar{0.236} & {0.090} & {0.617} & \underbar{0.931} & \underbar{0.989} \\
& VGLD-NK-TCI(Ours) & \textbf{0.241} & \textbf{0.545} & \textbf{0.433} & \textbf{0.189} & \textbf{0.067} & \textbf{0.779} & \textbf{0.967} & \textbf{0.993} \\
\hline

\multirow{6}{*}{DAV1-vits\cite{yang2024depth}}
& Least Squares & 0.196 & 0.416 & 0.341 & 0.282 & 0.061 & 0.875 & 0.963 & 0.981 \\

& Levenberg Marquardt & 0.151 & 0.440 & 0.234 & 0.108 & 0.029 & 0.957 & 0.989 & 0.995 \\
\cline{2-10} 
& RSA-NK\cite{zeng2024rsa}            & {0.290} & \underbar{0.563} & {0.571} & {0.250} & {0.092} & {0.640} & {0.899} & {0.969} \\
& VGLD-NK-T(Ours)   & {0.281} & {0.583} & {0.532} & {0.214} & {0.078} & {0.711} & {0.945} & {0.987} \\
& VGLD-NK-I(Ours)   & \underbar{0.250} & {0.573} & \underbar{0.443} & \underbar{0.194} & \underbar{0.070} & \underbar{0.764} & \underbar{0.965} & \underbar{0.991} \\
& VGLD-NK-TCI(Ours) & \textbf{0.241} & \textbf{0.545} & \textbf{0.433} & \textbf{0.189} & \textbf{0.067} & \textbf{0.779} & \textbf{0.967} & \textbf{0.993} \\
\bottomrule
\end{tabular}
}
\caption{Zero-shot generalization to SUN-RGBD (Indoor). Best results are in \textbf{bold}, second best are \underline{underlined}.}
\label{tab:3}
\end{table*}

\begin{table*}[h]
\centering
\setlength{\tabcolsep}{1mm} 
{\fontsize{9}{11}\selectfont
\begin{tabular}{l|l|ccccc|ccc}
\hline
\multirow{2}{*}{\textbf{RDE Model}} & \multirow{2}{*}{\textbf{Method}} & \multicolumn{5}{c|}{\textbf{Lower is better}} & \multicolumn{3}{c}{\textbf{Higher is better}} \\  
& & \textbf{Abs Rel $\downarrow$} & \textbf{sq\_rel $\downarrow$} & \textbf{RMSE $\downarrow$} & \textbf{$\text{RMSE}_\text{log}$ $\downarrow$} & \textbf{log$_{10}$ $\downarrow$} & \textbf{D1 $\uparrow$} & \textbf{D2 $\uparrow$} & \textbf{D3 $\uparrow$} \\
\hline

\multirow{6}{*}{MiDas-1\cite{birkl2023midas}} 
& Least Squares & 0.123 & 0.070 & 0.364 & 0.357 & 0.069 & 0.868 & 0.959 & 0.978 \\

& Levenberg Marquardt & 0.070 & 0.029 & 0.241 & 0.095 & 0.029 & 0.952 & 0.991 & 0.998 \\
\cline{2-10} 
& RSA-NK\cite{zeng2024rsa}             & {0.219} & \underbar{0.218} & {0.667} & {0.246} & {0.096} & {0.612} & {0.882} & {0.964} \\
& VGLD-NK-T (Ours)   & {0.251} & {0.385} & {0.683} & {0.240} & {0.094} & {0.622} & {0.898} & {0.969} \\
& VGLD-NK-I (Ours)   & \textbf{0.188} & \underbar{0.138} & \textbf{0.544} & \textbf{0.208} & \textbf{0.079} & \textbf{0.696} & \textbf{0.943} & \textbf{0.982} \\
& VGLD-NK-TCI (Ours) & \underbar{0.212} & {0.281} & \underbar{0.623} & \underbar{0.228} & \underbar{0.088} & \underbar{0.638} & \underbar{0.930} & \underbar{0.978} \\

\hline

\multirow{5}{*}{MiDas-2\cite{ranftl2020towards}}
& Least Squares & 0.133 & 0.080 & 0.394 & 0.345 & 0.071 & 0.846 & 0.954 & 0.977 \\

& Levenberg Marquardt & 0.086 & 0.039 & 0.285 & 0.114 & 0.036 & 0.929 & 0.988 & 0.996 \\
\cline{2-10} 
& VGLD-NK-T (Ours)   & {0.243} & \underbar{0.359} & {0.737} & \underbar{0.264} & \textbf{0.100} & \textbf{0.585} & {0.877} & {0.964} \\
& VGLD-NK-I (Ours)   & \underbar{0.235} & \underbar{0.201} & \underbar{0.722} & {0.294} & \underbar{0.115} & {0.460} & \underbar{0.849} & \underbar{0.975} \\
& VGLD-NK-TCI (Ours) & \textbf{0.227} & {0.371} & \textbf{0.690} & \textbf{0.262} & \textbf{0.100} & \underbar{0.570} & \textbf{0.894} & \textbf{0.979} \\

\hline

\multirow{5}{*}{DAV2-vits\cite{yang2024depthv2}}
& Least Squares & 0.123 & 0.068 & 0.361 & 0.361 & 0.069 & 0.872 & 0.960 & 0.978 \\

& Levenberg Marquardt & 0.066 & 0.024 & 0.226 & 0.092 & 0.028 & 0.958 & 0.993 & 0.998 \\
\cline{2-10} 
& VGLD-NK-T (Ours)   & {0.228} & \underbar{0.300} & {0.673} & \underbar{0.246} & \underbar{0.096} & \underbar{0.593} & {0.891} & {0.981} \\
& VGLD-NK-I (Ours)   & \underbar{0.212} & \textbf{0.169} & \underbar{0.663} & {0.259} & {0.103} & {0.514} & \underbar{0.899} & \underbar{0.989} \\
& VGLD-NK-TCI (Ours) & \textbf{0.196} & {0.487} & \textbf{0.610} & \textbf{0.208} & \textbf{0.082} & \textbf{0.678} & \textbf{0.952} & \textbf{0.990} \\

\hline

\multirow{6}{*}{DAV1-vits\cite{yang2024depth}}
& Least Squares & 0.118 & 0.063 & 0.345 & 0.344 & 0.066 & 0.875 & 0.961 & 0.979 \\

& Levenberg Marquardt & 0.056 & 0.020 & 0.203 & 0.081 & 0.024 & 0.970 & 0.994 & 0.999 \\
\cline{2-10} 
& RSA-NK\cite{zeng2024rsa}             & {0.216} & \underbar{0.283} & {0.679} & {0.249} & {0.098} & {0.608} & {0.873} & {0.964} \\
& VGLD-NK-T (Ours)   & {0.211} & {0.711} & {0.627} & \underbar{0.215} & \underbar{0.084} & \textbf{0.683} & {0.927} & {0.983} \\
& VGLD-NK-I (Ours)   & \textbf{0.193} & \textbf{0.200} & \textbf{0.597} & {0.220} & {0.087} & {0.619} & \underbar{0.950} & \textbf{0.994} \\
& VGLD-NK-TCI (Ours) & \underbar{0.196} & {0.487} & \underbar{0.610} & \textbf{0.208} & \textbf{0.082} &\underbar {0.678} & \textbf{0.952} & \underbar{0.990} \\

\bottomrule
\end{tabular}
}
\caption{Zero-shot generalization to DIML Indoor. Best results are in \textbf{bold}, second best are \underline{underlined}.}
\label{tab:4}
\end{table*}

\begin{table*}[h]
\centering
\setlength{\tabcolsep}{1mm} 
{\fontsize{9}{11}\selectfont
\begin{tabular}{l|l|ccccc|ccc}
\hline
\multirow{2}{*}{\textbf{RDE Model}} & \multirow{2}{*}{\textbf{Method}} & \multicolumn{5}{c|}{\textbf{Lower is better}} & \multicolumn{3}{c}{\textbf{Higher is better}} \\  
& & \textbf{Abs Rel $\downarrow$} & \textbf{sq\_rel $\downarrow$} & \textbf{RMSE $\downarrow$} & \textbf{$\text{RMSE}_\text{log}$ $\downarrow$} & \textbf{log$_{10}$ $\downarrow$} & \textbf{D1 $\uparrow$} & \textbf{D2 $\uparrow$} & \textbf{D3 $\uparrow$} \\
\hline

\multirow{6}{*}{MiDas-1\cite{birkl2023midas}} 
& Least Squares & 0.319 & 2.265 & 7.252 & 1.936 & 0.301 & 0.409 & 0.844 & 0.920 \\
& Levenberg Marquardt & 0.201 & 1.231 & 5.411 & 0.223 & 0.079 & 0.673 & 0.960 & 0.991 \\
\cline{2-10}
& RSA-NK\cite{zeng2024rsa} & {0.223} & {-} & {19.342} & {0.325} & {-} & {0.631} & \textbf{0.903} & \textbf{0.966} \\
& VGLD-NK-T (Ours) & {0.215} & {2.519} & {10.467} & {0.320} & {0.102} & {0.630} & {0.851} & {0.935} \\
& VGLD-NK-I (Ours) & \underline{0.212} & \textbf{2.409} & \textbf{10.061} & \textbf{0.311} & \underline{0.101} & {0.633} & {0.851} & {0.935} \\
& VGLD-NK-TCI (Ours) & \textbf{0.209} & \underline{2.517} & \underline{10.446} & \underline{0.319} & \textbf{0.100} & \textbf{0.659} & \underline{0.862} & \underline{0.941} \\

\hline

\multirow{5}{*}{MiDas-2\cite{ranftl2020towards}}
& Least Squares & 0.328 & 2.447 & 7.490 & 1.902 & 0.298 & 0.407 & 0.828 & 0.914 \\
& Levenberg Marquardt & 0.232 & 1.557 & 5.985 & 0.253 & 0.090 & 0.609 & 0.934 & 0.985  \\
\cline{2-10}
& VGLD-NK-T (Ours) & {0.232} & {2.625}& {14.324} & {0.326} & {0.112}  & {0.603} & {0.841} & {0.936} \\
& VGLD-NK-I (Ours) & \underline{0.220} & \underline{2.526} & \underline{12.235} & \underline{0.321} & \underline{0.106}  & \underline{0.642} & \underline{0.865} & {0.947} \\
& VGLD-NK-TCI (Ours) & \textbf{0.212} & \textbf{2.521} & \textbf{10.032} & \textbf{0.311} & \textbf{0.102}  & \textbf{0.659} & \textbf{0.881} & \textbf{0.954} \\

\hline

\multirow{5}{*}{DAV2-vits\cite{yang2024depthv2}}
& Least Squares & 0.318 & 2.239 & 7.205 & 1.937 & 0.300 & 0.410 & 0.847 & 0.920 \\
& Levenberg Marquardt & 0.173 & 1.027 & 4.988 & 0.200 & 0.069 & 0.757 & 0.974 & 0.992 \\
\cline{2-10}
& VGLD-NK-T (Ours) & {0.221} & {3.125} & {8.769} & {0.252} & {0.085} & {0.675} & {0.927} & {0.977} \\
& VGLD-NK-I (Ours) & \underline{0.185} & \underline{2.848} & \underline{8.344} & \textbf{0.232} & \textbf{0.074} & \underline{0.746} & \underline{0.929} & \underline{0.980} \\
& VGLD-NK-TCI (Ours) & \textbf{0.176} & \textbf{2.002} & \textbf{7.925} & \underline{0.238} & \underline{0.075} & \textbf{0.748} & \textbf{0.942} & \textbf{0.981} \\

\hline

\multirow{6}{*}{DAV1-vits\cite{yang2024depth}}
& Least Squares & 0.316 & 2.223 & 7.182 & 1.932 & 0.299 & 0.411 & 0.850 & 0.920 \\
& Levenberg Marquardt & 0.156 & 0.929 & 4.766 & 0.185 & 0.062 & 0.817 & 0.977 & 0.991  \\
\cline{2-10}
& RSA-NK\cite{zeng2024rsa} & {0.207} & {-} & {19.715} & {0.303} & {-} & {0.642} & {0.903} & \textbf{0.976} \\
& VGLD-NK-T (Ours) & {0.210} & {2.598} & {13.432} & {0.318} & {0.108} & {0.708} & {0.913} & {0.970} \\
& VGLD-NK-I (Ours) & \underline{0.192} & \underline{2.557} & \underline{9.275} & \underline{0.258} & \underline{0.081} & \underline{0.732} & \underline{0.922} & \underline{0.975} \\
& VGLD-NK-TCI (Ours) & \textbf{0.186} & \textbf{2.403} & \textbf{8.984} & \textbf{0.246} & \textbf{0.079} & \textbf{0.742} & \textbf{0.932} & \underline{0.975} \\
\bottomrule
\end{tabular}
}
\caption{Zero-shot generalization to DDAD  (Outdoor). Best results are in \textbf{bold}, second best are \underline{underlined}.}
\label{tab:10}
\end{table*}

\begin{figure*}[h]
\scriptsize
\centering
\setlength{\tabcolsep}{6pt}  
\includegraphics[width=0.7\textwidth]{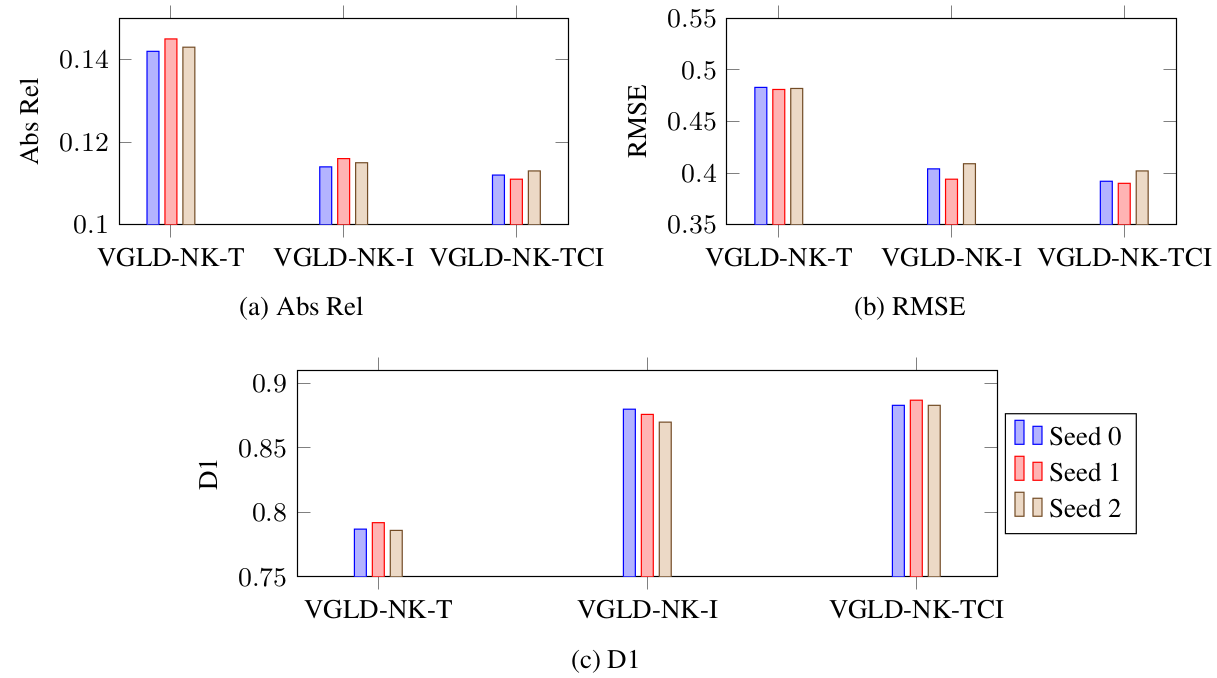}
\caption{Error bars showing performance variations across different random seeds (0, 1, 2) for Abs Rel, RMSE, and D1 metrics. Each group of bars corresponds to a specific variant of the VGLD model.}
\label{fig:diff_seed}
\end{figure*}

\begin{table*}[h]
\centering
\setlength{\tabcolsep}{3mm} 
{\fontsize{10}{12}\selectfont
\begin{tabular}{|c|p{13.9cm}|}
\hline
\textbf{Idx} & \textbf{Prompts} \\ 
\hline
1 &  Summarize the image in one sentence. \\
2 &  Summarize the image in one sentence, focusing mainly on the proximity relationships of the objects. \\
3 &  Describe the image in one sentence from near to far, focusing on the absolute positions of objects, with no more than 8 categories. \\
4 &  Describe the image in one sentence from near to far, focusing on the objects' relative positions, with no more than 8 categories. \\
5 &  Summarize the image in one sentence, describing the overall spatial layout of the image. \\
6 &  Summarize the image in one sentence, describing the overall distance relationships in the image. \\
\hline
\end{tabular}
}
\caption{Prompts for Natural Text Generation. We utilize two LLaVA models(\textit{llava-v1.6-vicuna-7b} and \textit{llava-v1.6-mistral-7b}), each generating 6 textual descriptions per image, resulting in a total of 12 diverse descriptions for each image.}
\label{tab:prompt text}
\end{table*}

\end{document}